\documentclass{article}

\usepackage[preprint]{neurips_2026}

\usepackage{booktabs}       %
\usepackage{longtable}
\usepackage{makecell}       %

\usepackage[utf8]{inputenc} %
\usepackage[T1]{fontenc}    %
\usepackage{hyperref}       %
\usepackage{url}            %
\usepackage{amsfonts}       %
\usepackage{nicefrac}       %
\usepackage{microtype}      %
\usepackage[table]{xcolor}
\usepackage{soul}           %
\usepackage{float}          %
\usepackage{tipa}           %

\usepackage[pdftex]{graphicx}
\usepackage{listliketab}
\usepackage{hyperref}
\hypersetup{
    colorlinks=true,
    linkcolor=blue,
    filecolor=magenta,
    urlcolor=blue,
    citecolor=blue
    }
\usepackage{amsmath}
\usepackage{ulem} %
\usepackage{cleveref} %
\usepackage{todonotes}
\usepackage[section]{placeins}
\usepackage{listings}

\usepackage[frozencache=true,cachedir=minted-cache]{minted}

\setminted{
  frame=none,
  bgcolor=gray!20,
  fontsize=\footnotesize,
  linenos,
  breaklines=true
} %

\usepackage{subcaption} %

\usepackage{float}
\floatstyle{plaintop}
\restylefloat{table}
\usepackage[tableposition=top]{caption}

\title{LibriBrain100: One Hundred Hours of Broad and Deep MEG Data for Neural Speech Decoding at Scale}

\author{%
Francesco Mantegna$^{1}$ 
\quad Dulhan Jayalath$^{1}$ 
\quad Gereon Elvers$^{1}$
\quad Tasha Kim$^{1}$
\AND Benjamin Ballyk$^{1}$ 
\quad Alex Fung$^{1,2}$ 
\quad SungJun Cho$^{1,3}$ 
\quad Teyun Kwon$^{1}$ 
\AND Luisa Kurth$^{1}$ 
\quad Miran \"{O}zdogan$^{1}$ 
\quad Gilad Landau$^{1}$ 
\quad Pratik Somaiya$^{1}$ 
\AND Natalie Voets$^{2}$ 
\quad Mark Woolrich$^{3}$ 
\quad Oiwi Parker Jones$^{1}$\\ \\
$^1$PNPL\includegraphics[height=2.2ex]{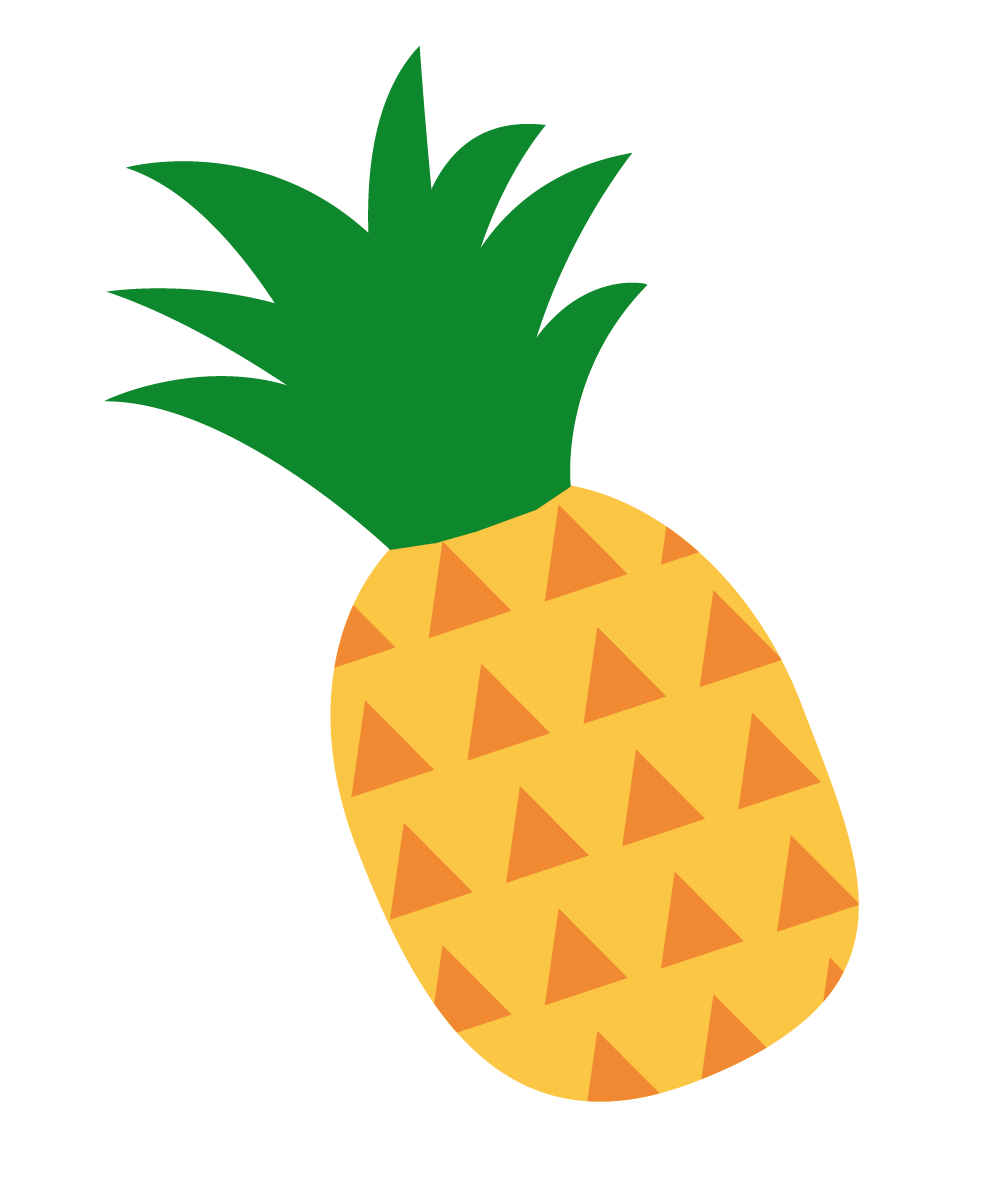}, Department of Engineering Science, University of Oxford, UK \\ \quad 
$^2$FMRIB, Oxford Centre for Integrative Neuroimaging, University of Oxford, UK \\
$^3$OHBA, Oxford Centre for Integrative Neuroimaging, University of Oxford, UK \\
\\ \texttt{\{francesco, oiwi\}@robots.ox.ac.uk}
}

\begin{document}

\maketitle
\begin{abstract}
We introduce LibriBrain100, a large-scale MEG dataset for speech decoding designed from the ground up for reproducible, standardised evaluation.
LibriBrain100 more than doubles the size of the original LibriBrain release, resulting in \textbf{over 100 hours of high-quality MEG} acquired while subjects listened to naturalistic continuous speech. 
With \textbf{$\sim$80 hours from a single subject}, LibriBrain100 sets a new record for deep, within-subject neural data (8$\times$ more than the next comparable dataset and roughly 80$\times$ more than other datasets). 
To demonstrate the payoff of this depth-first design, we evaluate on a word-classification benchmark---an increasingly well-established stepping stone towards the open challenge of non-invasive brain-to-text decoding. 
Using an existing decoding model, we achieve state-of-the-art performance---validating both the quality of the recordings and the value of within-subject data at scale.
Because collecting 80 hours of data per user is impractical for real-world applications, we also collected \textbf{$\sim$40 minutes of additional data from each of 32 subjects}. 
Using the same word-classification benchmark, we demonstrate the value of broad multi-subject data: supervised fine-tuning of a pre-trained model can substantially compensate for limited per-subject data.
We provide \textbf{standard train, validation, and test splits, all reproducible through an open-sourced Python library} that supports easy downloading, optional preprocessing, and data loading for common deep learning frameworks.
In addition, the dataset and evaluation infrastructure are being released alongside an open machine-learning competition with a public leaderboard for standardised benchmarking.
Ultimately, our hope is that LibriBrain100 will accelerate progress towards practical non-invasive brain-computer interfaces, capable of restoring communication to people living with severe paralysis. 

\end{abstract}

\section{Introduction}\label{sec:intro}

Deep learning had a transformational effect on computer vision once ImageNet gave it a benchmark: data at scale and standard evaluation protocols. Neural speech decoding is at a similar inflection point --- but with an added complexity. In invasive settings, systems decoding from surgically implanted electrodes now outperform automatic speech recognition in some conditions, achieving word error rates below 5\% in paralysed patients \citep{card2024nejm}. But invasive recording requires brain surgery, limiting deployment to clinical trials and excluding the vast majority of people who might benefit. Non-invasive alternatives are therefore the real prize, and deep learning is beginning to show real gains there too \citep{tang2023, defossez2023, dascoli2025natcomm, jayalath2025unlocking}. Yet unlike computer vision, the field has lacked the shared infrastructure to know how much progress is real, what can be built on reliably, and how quickly to anticipate advances. %
Without standard benchmarks, shared data, and common evaluation protocols, it is difficult to know what works or by how much.

The LibriBrain dataset \citep{ozdogan2025libribrain} --- and its open ML competition \citep{landau2025competition} --- represented a significant step in the right direction. Taking the insight that within-subject data pushes results fastest \citep{dascoli2025natcomm}, LibriBrain provided the largest within-subject magnetoencephalography (MEG) dataset for speech decoding from brain recordings to date, together with standard splits and a curriculum of benchmark tasks. Building on the LibriBrain infrastructure, subsequent work has advanced the state of the art across speech detection, phoneme classification, word classification, keyword spotting, and even full brain-to-text decoding \citep{elvers2025elementary, elvers2026benchmarking, jayalath2025unlocking, jayalath2025meg-xl}.

But LibriBrain has its limitations. It covers only a single subject listening to a single genre (detective fiction) and so does not speak directly to cross-subject generalisation. This is of a critical importance for real-world BCIs, where per-user data collection must be minimal. The limited stimulus diversity inherent in a tight focus on the series of Sherlock Holmes books, leaves nuanced questions about phonetic and semantic decoding underexplored. Finally, even with its unprecedented scale of within-subject data, analyses of LibriBrain have shown no signs of plateauing~\citep[see, e.g.,][]{ozdogan2025libribrain}, suggesting that more data should yield even bigger gains.

LibriBrain100 addresses these limitations directly. It extends LibriBrain both in \textit{depth} and in \textit{breadth}. Along the depth axis, LibriBrain100 extends the within-subject data from $\sim$50 to $\sim$80 hours, completing the entire canon of Sherlock Holmes and adding new stimuli designed with both phonetics and semantics in mind: TIMIT~\citep{garofolo1993timit} and MOCHA-TIMIT \citep{wrench1999mocha_timit, wrench2000mocha_timit} are standard corpora in both ASR research \citep[e.g.,][]{mohamed2009deep, hinton2012asr, graves2013speech} and invasive speech BCIs \citep[e.g.,][]{mesgarani2014phonetic, cheung2016, anumanchipalli2019, makin2020, metzger2023neuroprosthesis}, which control for phoneme and phoneme-pair distributions and enable direct comparison with prior studies using the same stimuli; and podcast narratives spanning a wide range of semantic domains, which were previously used for fMRI-based semantic decoding \citep{tang2023} enabling cross-modal comparison. 
Along the breadth axis, LibriBrain100 extends data collection to 32 new subjects, each contributing $\sim$40 minutes of MEG, designed to support supervised fine-tuning and data-efficient cross-subject generalisation. 
Together, these corpora span the acoustic--semantic axis, positioning LibriBrain100 to address the questions of how MEG-based speech decoding may be driven by representations of sound versus meaning in the brain.

\begin{figure}[t!]
    \centering
    \includegraphics[width=\linewidth]{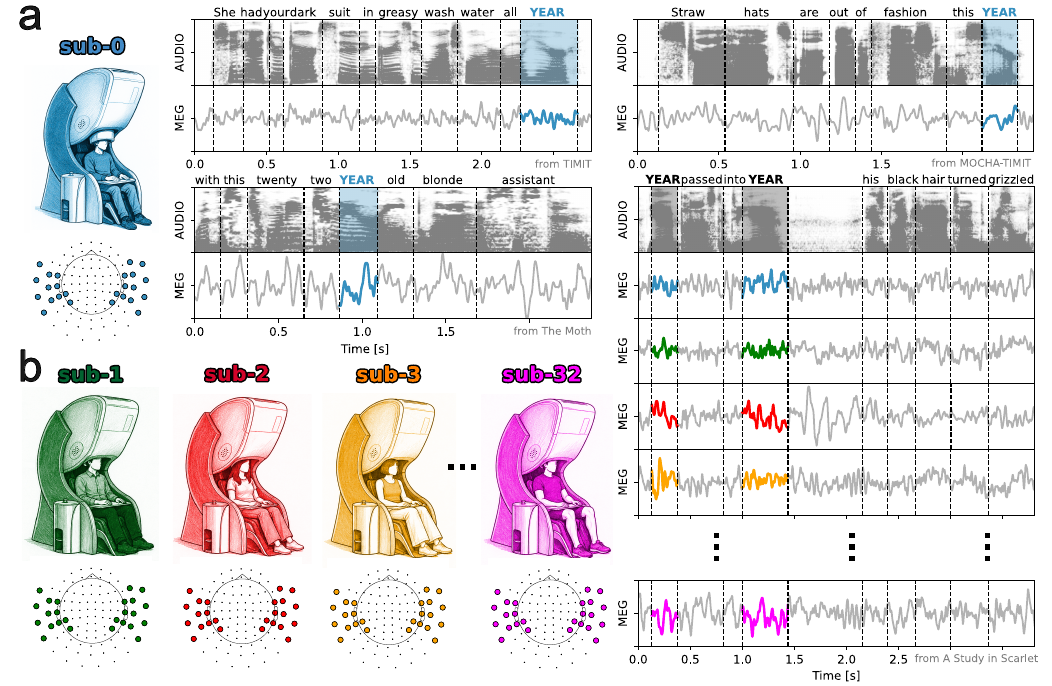}
    \caption{\textbf{Decoding words from LibriBrain100}. LibriBrain100 supports word-classification benchmarks across both deep within-subject data and broad cross-subject data. (\textbf{a}) \textit{Within-subject decoding}: example speech stimuli and MEG responses from subject 0 across multiple corpora. Target words are highlighted in the text, aligned to the acoustic spectrogram, and marked in the corresponding MEG time series. MEG time series were averaged across temporal sensors illustrated in the sensor layout. (\textbf{b}) \textit{Cross-subject decoding}: the same target-word classification task can be evaluated across the 32 additional subjects with limited per-subject data, enabling benchmarking for data-efficient generalisation.}
    \label{fig:task}
\end{figure}

On the evaluation side, LibriBrain100 advances the curriculum of tasks from speech detection and phoneme classification to word classification --- a task capable of utilising both phonetic and semantic representations in the brain, and that sits closer to the brain-to-text goal that motivates much of the field (Figure~\ref{fig:task}). We evaluate word classification in two settings: within-subject, targeting state-of-the-art performance by scaling up training data for subject~0; and cross-subject, focusing on data-efficient generalisation via supervised fine-tuning from a single high-data subject to new subjects with limited recordings. Standard splits are provided across all sub-datasets, and baseline results using MEG-XL \citep{jayalath2025meg-xl} are released to support reproducible community benchmarking.

LibriBrain100 makes the following contributions:
\begin{itemize}
    \item \textbf{A large-scale, stimulus-diverse within-subject MEG dataset}: ${\sim}80$ hours of recordings from subject~0 --- the largest within-subject speech decoding dataset to date --- completing the full Sherlock Holmes canon and adding TIMIT and MOCHA-TIMIT for phonetically controlled analyses and podcast narratives for semantic decoding --- enabling cross-modal comparison with the fMRI literature and direct comparison with influential invasive BCI studies
    \item \textbf{A multi-subject MEG dataset for cross-subject generalisation}: recordings from 32 new subjects (${\sim}40$ minutes each), designed to support supervised fine-tuning and data-efficient generalisation benchmarking
    \item \textbf{A word classification benchmark with standard splits and baselines}: an extended evaluation curriculum with train/validation/test splits across all sub-datasets, together with reproducible baseline results for within-subject and cross-subject settings using MEG-XL
    \item \textbf{Open data release and tooling}: raw and preprocessed data on HuggingFace, rich linguistic annotations spanning sub-lexical to lexical levels, and a Python library for streamlined deep learning integration --- all resources to support a planned open ML competition
\end{itemize}

\section{Related Work}\label{sec:related_work}
In recent decoding work on MEG acquired during naturalistic speech listening~\citep[e.g.,][]{defossez2023, dascoli2025natcomm, jayalath2025unlocking, jayalath2025bbl-icml, jayalath2025meg-xl}, a recurring set of large-scale MEG datasets have emerged \citep[also see][]{king2026neuralset}. 
The usual suspects include MOUS~\citep{schoffelen2019mous}, MEG-MASC~\citep{gwilliams2023megmasc}, \citet{armeni2022}, Le Petit Prince~\citep{dascoli2025natcomm}, and LibriBrain~\citep{ozdogan2025libribrain}. 
Table~\ref{tab:datasets} (Appendix~\ref{app:existing_datasets}) summarises these datasets along with LibriBrain100, listing details such as the language of stimuli and dimensions in which we can measure their size. 
In terms of `deep', within-subject data, LibriBrain100 is $\sim$8$\times$ bigger than the next non-LibriBrain dataset~\citep{armeni2022}, and $\sim$80$\times$ bigger than the rest. 

We can compare the existing datasets in terms of \textit{breadth}---the number of subjects---and \textit{depth}---the hours of data per subject. On the one hand, MOUS, MEG-MASC, and Le Petit Prince represent breadth-first designs, spanning 27--96 subjects but with under 2 hours per subject. Armeni and LibriBrain contrast depth-first designs, with 10 and 52.3 hours per subject respectively, but only 1--3 subjects. This distinction matters for decoding. Despite other datasets having greater total hours, \citet{dascoli2025natcomm} find word-classification performance to be strongest by a substantial margin for Armeni---the deepest dataset in their study, LibriBrain having not come out yet. When LibriBrain is included \citep[as in][]{jayalath2025unlocking}, decoding performance is even stronger with it than with Armeni, further underscoring the importance of within-subject depth.

The obvious trade-off for LibriBrain, which is the closest dataset to LibriBrain100, is that its depth comes at the cost of any breadth---the original LibriBrain release contained just one subject. LibriBrain100 addresses this limitation while continuing to prioritise depth. It deepens the single-subject recording to $\sim$80 hours while adding $\sim$40 minutes of data from each of 32 additional subjects---resulting in a hrs/subject range of 0.6--80.0. This reflects both a depth-first design and respectable provisions for studying generalisation to new subjects with more realistic amounts of data---of the six datasets in Table~\ref{tab:datasets}, LibriBrain100 has the 3rd highest number of subjects. 
The nature of the audio stimuli in these datasets is also interesting. Unlike LibriBrain, which focused only on readings of Sherlock Holmes (seven out of nine books), LibriBrain100's stimulus set---which includes TIMIT~\citep{garofolo1993timit}, MOCHA-TIMIT~\citep{wrench1999mocha_timit, wrench2000mocha_timit}, and 30 podcasts-from \textit{The Moth} (\url{https://themoth.org})--is designed to better control for sound and meaning-based brain activity.

\section{The LibriBrain100 Dataset}\label{sec:dataset}

    \subsection{Overview}\label{sec:overview}

The LibriBrain100 dataset contains over 100 hours of non-invasive magnetoencephalography (MEG) recordings, which were acquired while subjects listened to connected speech (Table~\ref{tab:data_splits}). 
In addition, the dataset includes paired event files with time-locked annotations for linguistic events (e.g.~speech/non-speech, phonemes, words). These annotations were designed for use as labels for decoding tasks such as word classification (Section \ref{sec:experiments}). 
The MEG data are being released in both raw and preprocessed/serialised formats, where both may be thought of as matrices with 306 sensor dimensions $\times$ $T$ time samples. %
MEG recordings were originally sampled at 1 kHz but then downsampled to 250 Hz during preprocessing to preserve oscillations into the high-gamma range (70–125 Hz). 
Consequently, each sample $t \in [1, T]$ represents 1 ms in the raw data and 4 ms in the preprocessed data.

        \begin{table}[!htbp]
\centering
\caption{\textbf{LibriBrain100 data sources and standard splits}. The dataset comprises 
a \textit{deep} single-subject component ($\sim$80 hours) and a \textit{broad} 
multi-subject component ($\sim$40 minutes per subject across 32 subjects). 
All durations are shown in hours and rounded to one decimal place; totals are computed before rounding.
For a visual summary of the dataset, see Figure~\ref{fig:broad+deep} (Appendix~\ref{app:dataset_proportions}).
}
\label{tab:data_splits}
\small
\begin{tabular}{llrrrrr}
\toprule
Component & Source & \# Subj & All Data & Train & Val & Test \\
\midrule
\textit{Deep} & & & & & & \\
\quad Subject 0 & Sherlock (all books)  & 1 & 68.1 & 67.3 & 0.4 & 0.4 \\
               & TIMIT                  & 1 &  5.3 &  4.9 & 0.2 & 0.2 \\
               & MOCHA-TIMIT            & 1 &  1.1 &  0.8 & 0.2 & 0.2 \\
               & The Moth (30 podcasts) & 1 &  6.0 &  5.7 & 0.2 & 0.2 \\
\quad Subtotal (hours) &                        & 1 & 80.5 & 78.6 & 1.0 & 0.9 \\
\midrule
\textit{Broad} & & & & & & \\
\quad Subjects 1--32 & Sherlock (2 chapters) & 32 & 23.7 & -- & 11.5 & 12.2 \\
\quad              & ($\sim$44 minutes/subject) &   &       &   &       &       \\
\midrule
\textbf{Total (hours)} &  & \textbf{33} & \textbf{104.2} & \textbf{78.6} & \textbf{12.5} & \textbf{13.1} \\
\bottomrule
\end{tabular}
\end{table}

    \subsection{Data Collection and Structure}\label{sec:collection+structure}

All MEG data were collected at the Oxford Centre for
Human Brain Activity (OHBA) on a MEGIN TRIUX™ Neo system with 306 sensors known as superconducting quantum interference devices (SQUIDs)~\citep{hamalainen1993}. The SQUIDs in this system partition into 102 magnetometers and 204 gradiometers. %
In a magnetically shielded room, these sensors are sufficiently sensitive to detect electromagnetic fields generated by cortical activity throughout the whole brain~\citep{benar2021}. %
The present release includes data from 33 subjects, all of whom were proficient English speakers and provided informed consent for pseudonymised data to be shared; this study was approved by the University of Oxford Medical Sciences Interdivisional Research Ethics Committee (R90053/RE003). 

In terms of structure, the majority of the data ($\sim$80 hours) comes from a single subject, referred to as `subject zero' (\texttt{sub-0}). This was a deliberate choice, as hours of data per subject is known to drive performance more than total number of hours across subjects~\citep{dascoli2025natcomm}. At the same time, for downstream applications where it would be preferable not to require such large data from every new subject, we also include more practical amounts of data ($\sim$20 minutes $\times$ 2 sessions per subject) from a cohort of 32 additional subjects (\texttt{sub-1} to \texttt{sub-32}). In total, the dataset includes more than 100 hours of MEG data with annotations generated from the audio stimuli that the subjects listened to. %

The audio stimuli include LibriVox recordings for every book in the canon of Sherlock Holmes~\citep{doyle1887study, doyle1890sign, doyle1892adventures, doyle1893memoirs, doyle1902hound, doyle1905return, doyle1915valley, doyle1917lastbow, doyle1927casebook}. While the first seven of these books ($\sim$50 hours) featured in the original LibriBrain dataset~\citep{ozdogan2025libribrain}, completing the canon here was not only completionist but aimed to maximise data quantity while minimising variability. For example, all but one book were read by the same narrator and both narrators selected to have similar Southern British accents (see Appendix \ref{sec:stimuli} for more details). Our aim in limiting the variability of these stimuli was pragmatic. By analogy, if you wanted to train automatic speech recognition (ASR) from scratch on about 68 hours of audio recordings, it would be easier to concentrate on just one or two speakers than attempt to model every accent or dialect. 

To complement the Sherlock Holmes audiobooks, we looked to the literature for inspiration. For better phonetic coverage, we used TIMIT~\citep{garofolo1993timit} and MOCHA-TIMIT~\citep{wrench1999mocha_timit, wrench2000mocha_timit}. 
Not only has TIMIT long been a cornerstone for the development of ASR technology~\citep[e.g.,][]{graves2013speech}, but it has played a foundational role in our understanding of speech comprehension in the brain~\citep[e.g.,][]{mesgarani2014phonetic, cheung2016}. 
One feature of TIMIT is that the sentences in it were designed to balance the distribution of phonemes and phoneme-bigrams. Unlike Sherlock, TIMIT also contains audio recordings from 630 speakers across 8 dialect groups in the USA; 
MOCHA-TIMIT adds male and female speakers from the UK (see Appendix \ref{sec:stimuli}).

For semantic coverage, we looked to a landmark semantic decoding study~\citep{tang2023} which used multiple short ($\sim$12 minute) podcast stories to cover a broad range of topics. We include 30 podcast stories from \textit{The Moth} (for details, see Appendix \ref{sec:stimuli}).

For reproducible evaluation, the data are split into standard training, validation, and testing sets. 
Following both~\citet{ozdogan2025libribrain} and~\citet{landau2025competition}, sessions 11 and 12---which correspond to chapters 11 and 12---from the first Sherlock book are held out for validation and testing, respectively. 
Sessions 13 and 14, which were previously held-out for competition evaluation~\citep{landau2025competition}, have now been made public and can be used for training, together with the rest of the Sherlock data. 
As there is a standard split for TIMIT, we use that in the MEG data for comparability.
Concretely, we use the 24-speaker core test set, which consists of 2 male and 1 female speaker from each dialect region, totalling 192 utterances (24 speakers $\times$ 8 utterances which exclude the shared \texttt{SA1} and \texttt{SA2} sentences). 
For validation, we follow the standard Kaldi TIMIT recipe and use its 50-speaker development set, again excluding the shared \texttt{SA1} and \texttt{SA2} sentences.
Although surgical decoding studies have used MOCHA-TIMIT~\citep[e.g.,][]{anumanchipalli2019,makin2020}, we are unaware of a precedent for a standard split; therefore, we created our own split. To do this, we note that MOCHA-TIMIT is organised into four sets, A--D, where sentences repeat between sets A and D and between sets B and C. We use A and D for training. For validation and testing, we split the sentences in sets B and C into equally sized subsets of unique sentences. This results in unique sentences for train, validation, and test sets, minimising the risk of information leakage due to sentence type.
Table \ref{tab:data_splits} summarises the data and splits.

    \subsection{Data Formats, Access, and Supporting Python Library}\label{sec:formats+access}

We are releasing LibriBrain100 in two formats. 
First, we are releasing the raw data (with no preprocessing) in BIDS format~\citep{niso2018megbids}. BIDS structures are directories that follow guidelines on file naming and organisation, with MEG data in FIF format. 
Technically, LibriBrain100 is a collection of BIDS datasets, one for each Sherlock book, TIMIT, MOCHA-TIMIT, and The Moth podcasts.
Second, we are releasing the data in a ready serialised format. These data have been minimally preprocessed (i.e., corrected for head movement, filtered, downsampled) and then converted to float32 HDF5 format, for easy machine learning. 

The MEG data in both FIF and HDF5 files can be thought of as a 306 channel $\times$ $T$ time-sample matrix, with time sampled at 1,000 Hz for the raw data or 250 Hz for the serialised data. 
Each FIF or HDF5 file is paired with an annotation file in TSV format. The annotation files contain time-stamps for linguistic events like word onsets, which can be used to define labels for supervised learning. 

The recommended way to access the data is through our custom Python library, \texttt{pnpl} (\verb|pip install pnpl|). 
The library provides task-driven \texttt{torch.utils.data.Dataset} classes whose constructors return ready-to-iterate windows of MEG, time-locked to events such as word or phoneme onsets:
\begin{minted}{python}
from pnpl.datasets import LibriBrain100
from pnpl.tasks import WordClassification

ds = LibriBrain100(
    data_path="./data/LibriBrain100",
    task=WordClassification(tmin=0.2, tmax=0.6),
    partition="train",
)

x, y = ds[0]   # x: (channels, time), y: integer word class id
\end{minted}
If the requested files are not at \texttt{data\_path}, the library downloads them on demand from Hugging Face (the \texttt{LibriBrain100} loader transparently fetches each record from whichever underlying repository owns it). Given the size of the full dataset ($\sim$104 hours of MEG, hundreds of GB even after serialisation), the constructor exposes selectors so users can scope the download to the part they care about:

\begin{minted}{python}
# Subject 0 only -- the deep component (~80 hours):
ds = LibriBrain100(data_path=..., task=..., partition="train",
                   subjects="deep")

# Only TIMIT, only on subject 0:
ds = LibriBrain100(data_path=..., task=...,
                   subjects="deep", corpus="timit")

# Multiple corpora at once, still only on subject 0:
ds = LibriBrain100(data_path=..., task=...,
                   subjects="deep", corpus=["timit", "mocha"])
\end{minted}

The \texttt{subjects=} argument accepts \texttt{"all"} (default), \texttt{"deep"} (subject 0), \texttt{"broad"} (subjects 1--32), an integer, a single subject id (\texttt{"0"} or \texttt{"sub-0"}), or any iterable of those. The \texttt{corpus=} argument accepts \texttt{"all"} (default), \texttt{"sherlock"}, \texttt{"timit"}, \texttt{"mocha"}, \texttt{"podcasts"}, or a list of those (the canonical token for ``MOCHA-TIMIT'' is \texttt{"mocha"} but aliases like \texttt{"mocha-timit"} or \texttt{"the\_moth"} are accepted). As the broad component (subjects 1--32) was only collected with the Sherlock stimuli, the loader rejects combinations like \texttt{subjects="broad", corpus="timit"} up front rather than silently returning an empty or incomplete dataset. Per-task wrappers (\texttt{LibriBrain100Speech}, \texttt{LibriBrain100Phoneme}, \texttt{LibriBrain100Word}) are also available for convenience. 
For finer control, users can pass a list of explicit \texttt{include\_run\_keys} or \texttt{exclude\_run\_keys} as \texttt{(subject, session, task, run)} tuples, mirroring the run-key convention from the original LibriBrain release.

While we recommend accessing the data through the Python library, with its high-level API which does not require an understanding of the underlying repository structure, the data are also available for direct download from two Hugging Face repositories: the original LibriBrain repository
(\url{https://huggingface.co/datasets/pnpl/LibriBrain}) and a new LibriBrain2 repository (\url{https://huggingface.co/datasets/pnpl/LibriBrain2}). 
For philosophers and fans of category mistakes \citep{ryle1949concept}, we might point out that LibriBrain100 is not a third repository alongside LibriBrain and LibriBrain2, but rather their union --- as the University of Oxford is not a building alongside its colleges, departments, and other facilities, but rather the collection of them all.

\section{Decoding Experiments}\label{sec:experiments}
To evaluate LibriBrain100, we focus here on the task of word classification which is defined as follows. 
Given a multi-channel MEG recording $\mathbf{X} \in \mathbb{R}^{C \times T}$ with $C$ channels and $T$ time samples, the word classification task maps to a target word $y \in \mathcal{V}$ where $\mathcal{V}$ is a fixed vocabulary. 
Word classification has previously been studied in MEG using the most frequent 50 and 250 words~\citep{dascoli2025natcomm, jayalath2025unlocking, jayalath2025meg-xl} defined over various corpora and using top-10 balanced accuracy, computed by averaging per-class top-$10$ accuracy across the vocabulary $\mathcal{V}$ to account for imbalances in word frequency~\citep{zipf1949zipfslaw}. 
In the following experiments, we use a vocabulary of 50 words (see Appendix \ref{sec:target_words}). 

As a backbone, we use MEG-XL~\citep{jayalath2025meg-xl}. This pre-trained model has achieved state-of-the-art word classification results in cross-subject MEG decoding using unsupervised pre-training followed by supervised fine-tuning. 
MEG-XL draws on insights from LLMs to significantly improve downstream performance by extending the temporal context (5--300$\times$) of previous foundation models of the brain~\citep[cf.~][]{labram2024, avramidis2025biocodec, wang2024eegpt, xiao2025brainomni, wang2025cbramod}. 
MEG-XL has been pre-trained on approximately 300 hours from 800 subjects over multiple speech and non-speech datasets---i.e., MOUS~\citep{schoffelen2019mous}, Cam-CAN~\citep{shafto2014camcan}, and SMN4Lang~ \citep{wang2022smn4lang}---before being fine-tuned on LibriBrain~
\citep{ozdogan2025libribrain}. 
In this paper, we fine-tune the base MEG-XL checkpoint (\url{https://huggingface.co/pnpl/MEG-XL}) on LibriBrain100 for the following tasks.

\subsection{Task 1: From Many-to-One --- Improving Within-Subject Word Classification}\label{sec:within_subj_task}

\begin{figure}
    \centering
    \includegraphics[width=1.0\linewidth]{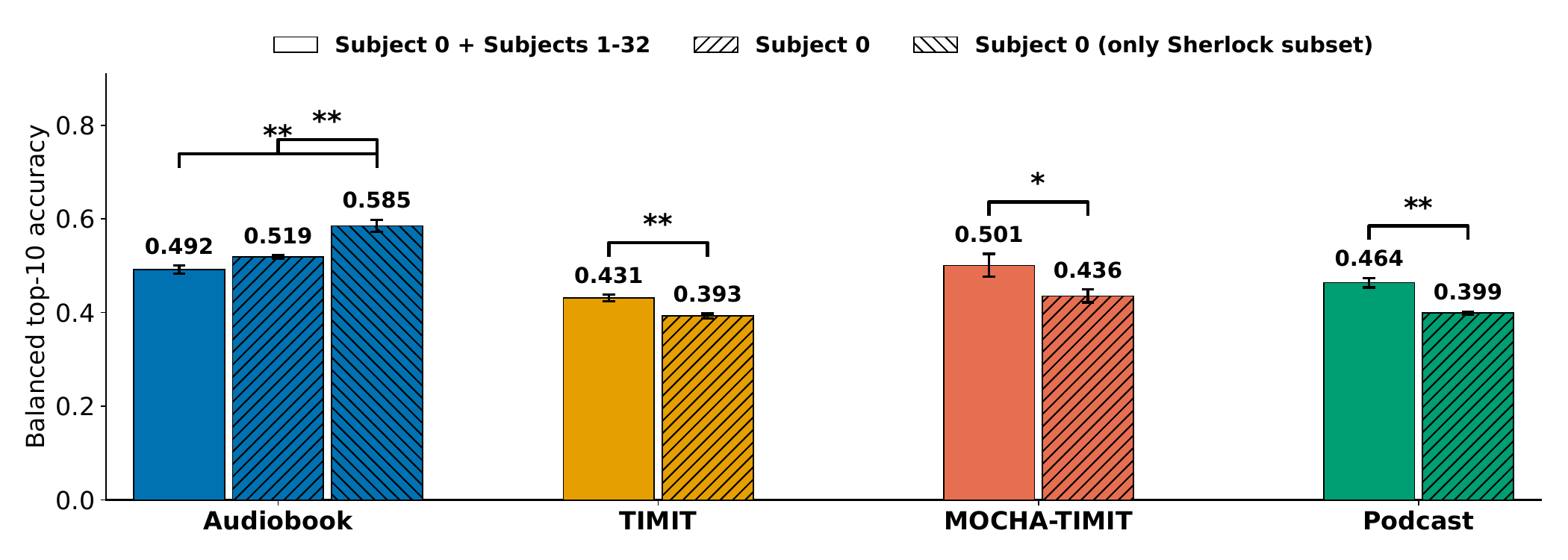}
    \caption{\textbf{Subject 0 performance on held-out test data}. Performance improves when training jointly with Subject 1-32 training data, except on the Sherlock subset. Here, training only on Sherlock training data leads to the best performance. Random chance accuracy is $0.2$. * indicates $p<.05$ and ** indicates $p<.01$ under Mann-Whitney U-tests.}
    \label{fig:deep}
\end{figure}

Figure~\ref{fig:deep} describes the results on the held-out test data of Subject 0 when fine-tuning MEG-XL in two settings. In the first, we fine-tune MEG-XL on all of the available training data in LibriBrain100; in the second, we exclude the training data of subjects 1--32. Training with the broad subject data appears to be helpful for generalisation to different speech distributions in the form of TIMIT, Mocha-TIMIT, and the podcast data. However, when evaluating on Sherlock, which forms the majority of the data, training without subjects 1-32 led to the best test performance, though only marginally. 
In general, even when we have a large amount of data for a single subject (Subject 0), additional data from 32 other subjects appears to improve downstream performance on the single subject (Subject 0).

    \subsection{Task 2: From One-to-Many --- Improving Between-Subject Word Classification}\label{sec:between_subj_sft_task}

\begin{figure}
    \centering
    \includegraphics[width=1.0\linewidth]{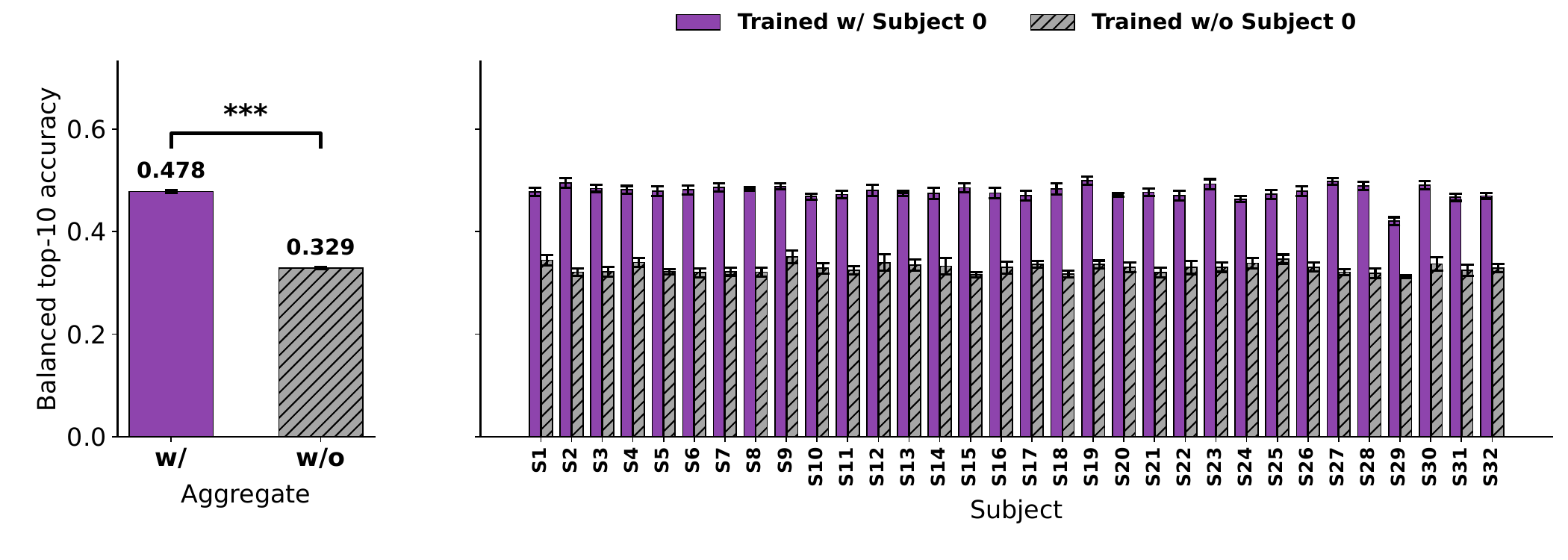}
    \caption{\textbf{Fine-tuning MEG-XL on subjects 1-32, with and without training alongside Subject 0}. We report performance on the Sherlock test set. Training jointly with Subject 0 significantly improves generalisation on the broad and shallow data of subjects 1-32. Random chance accuracy is $0.2$. *** indicates $p<.001$ under a Mann-Whitney U-test. The per-subject results in the right panel illustrate that difference is not driven by any specific subject, or proper subset of subjects, but is rather consistent across all 32 subjects.}
    \label{fig:breadth}
\end{figure}

In Figure~\ref{fig:breadth}, we assess generalisation to the held-out test data on subjects 1--32. Similar to the previous section, we fine-tune MEG-XL in two settings. Here, we analyse training on the multi-subject training data jointly with Subject 0's training data (``Training w/ Subject 0''), and also the effect of excluding the Subject 0 training data (``Training w/o Subject 0''). All subjects generalise similarly, and including Subject 0's training data improves generalisation across subjects by about 15 percentage points. 
This shows that a large amount of data from a single subject ($\sim$80 hours) can be used to improve performance on many individuals for whom we have $\sim$120$\times$ less data ($\sim$40 minutes each). 
In the next section, we investigate pushing this discrepancy further, reducing the amount of fine-tuning data needed for subjects 1--32 down to $\sim$10 minutes apiece. 

\subsection{Task 3: Exploring Efficiency --- Maintaining Improvements in Between-Subject Word Classification with Decreasing amounts of Supervised Fine-Tuning Data}\label{sec:efficiency_task}

Here we investigate the impact of reducing the available fine-tuning data for subjects 1--32, building on the results from Section~\ref{sec:between_subj_sft_task}. This is an important test for future applications, as even 40 minutes of brain recordings may be too much for patients to use a speech BCI. 
A speech brain--computer interface should generalise to clinical patients with minimal subject-specific training.

\begin{figure}
    \centering
    \includegraphics[width=1.0\linewidth]{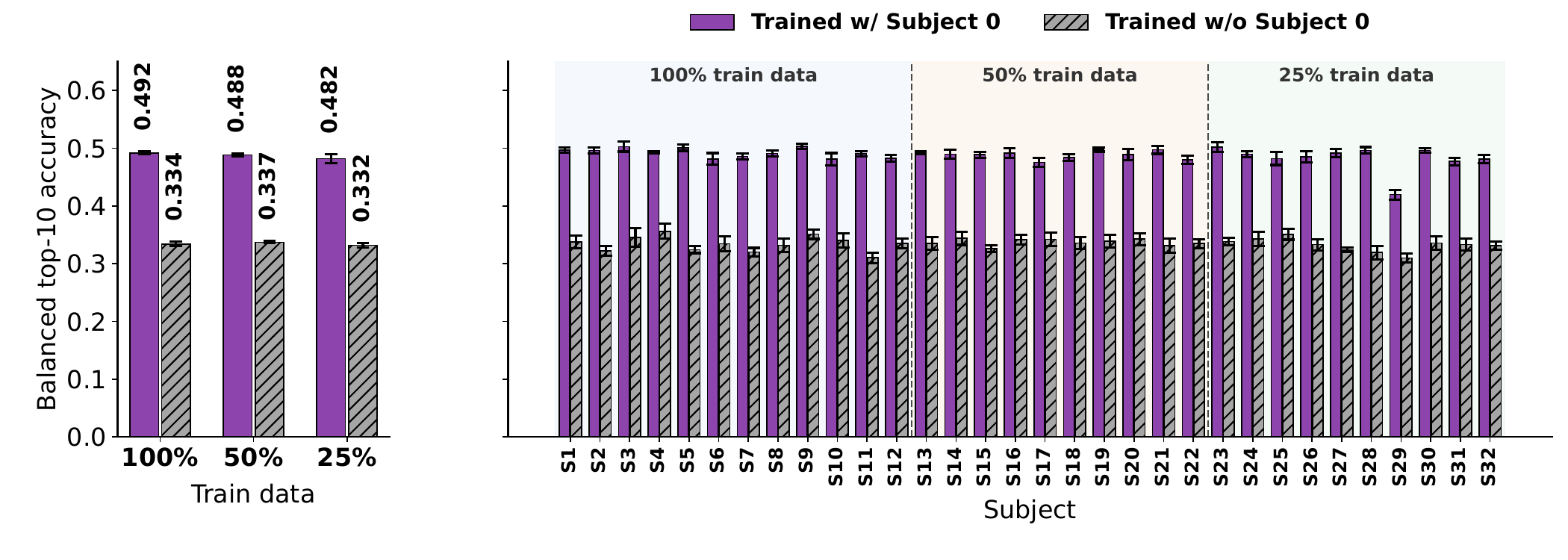}
    \caption{\textbf{Performance as a function of fine-tuning data size}. Generalisation remains robust, even with only 25\% training data, equivalent to about 10 minutes of brain recordings (i.e., a small enough amount of data to be interesting for clinical applications). The largest difference comes from including data from Subject 0. Random chance accuracy is $0.2$. None of the differences in performance under different training data percentages were significant under Mann-Whitney U-tests. }
    \label{fig:efficient}
\end{figure}

In Figure~\ref{fig:efficient}, we compare the use of 100\% of the available training data for 12 subjects, with 50\% for the next 10 subjects, and 25\% for the remaining 10 subjects. In all cases, we see minimal degradation on generalisation, suggesting that cross-subject transfer is responsible for most of the present decoding performance. 
Our strategy of pre-training on deep single-subject data and fine-tuning to new individuals appears to be sound, and may even be practical for future clinical applications --- though we emphasise that all LibriBrain100 data was acquired from healthy volunteers. 

\section{Discussion}\label{sec:discussion}

LibriBrain100 substantially extends the LibriBrain ecosystem in both scale and scope. With ${\sim}80$ hours from a single subject and ${\sim}40$ minutes from each of 32 additional subjects, it provides the largest within-subject MEG dataset for speech decoding to date while simultaneously opening the door to cross-subject generalisation research.

\paragraph{Limitations addressed.}
The original LibriBrain paper closed with a list of limitations. LibriBrain100 directly addresses several of them. First, the \textit{single-subject design}: the addition of 32 new subjects makes cross-subject generalisation a first-class benchmark task for the first time in this dataset series. Second, \textit{focused language content}: whereas LibriBrain drew exclusively from books in the Sherlock Holmes series, LibriBrain100 adds stimuli deliberately spanning the phonetic--semantic axis --- TIMIT and MOCHA-TIMIT for controlled phonetic coverage, and podcast narratives for broader semantic diversity~\citep{tang2023}. We expect these sub-datasets will be useful in their own right, enabling targeted studies of phonetic and semantic decoding independently of the collective LibriBrain100 datasets. Third, \textit{preprocessing}: the serialised HDF5 release lowers the barrier to entry for ML researchers by removing the need for domain-specific signal processing knowledge, while the raw BIDS release preserves full flexibility for neuroscientists who wish to apply their own pipelines.

\paragraph{Limitations remaining.}
Two limitations from the original LibriBrain list remain open. The first is the \textit{listening paradigm focus}. All data in LibriBrain100 were collected during passive listening to continuous speech. Listening provides a natural starting point for speech decoding, as alignments between stimulus and neural response are straightforward to derive, and signal-to-noise is comparatively high. However, it is not the paradigm that ultimately matters for BCIs, where users must attempt to produce or imagine speech. We have been actively collecting inner speech data in parallel and plan to release multiple inner speech datasets going forward. The second remaining limitation is \textit{brain-to-text}. We made a deliberate decision not to include a brain-to-text baseline in this release. Progress on the open-ended decoding problem has been rapid but uneven, and we judged that the field is not yet at a point where a single baseline result would be stable enough to anchor community benchmarking --- however, we will revisit this in future releases as methods mature.

\paragraph{Conclusion.}
LibriBrain100 follows the logic of ImageNet: the value of a shared benchmark compounds over time as more methods are trained and evaluated against it. The open ML competition, public leaderboard, and standardised Python library are all designed with cumulative progress in mind. We hope that LibriBrain100, like its predecessor, will serve not just as a dataset but as shared infrastructure --- lowering the cost of entry for ML researchers, enabling rigorous comparison across methods, and ultimately accelerating progress toward practical non-invasive brain-computer interfaces capable of restoring communication to people living with severe paralysis.

\begin{ack}

We are grateful to our anonymous reviewers for their time and attention, which helped to make this paper better, and to the NeurIPS Evaluations \& Datasets organisers for all of their efforts in making the track possible. 
We acknowledge the use of the University of Oxford Advanced Research Computing (ARC) facility (\url{http://dx.doi.org/10.5281/zenodo.22558}), Hartree Centre resources, and the NVIDIA Corporation for donating additional GPUs. PNPL is supported by the MRC (MR/X00757X/1), Royal Society (RG$\backslash$R1$\backslash$241267), NSF (2314493), NFRF (NFRFT-2022-00241), SSHRC (895-2023-1022), and ARIA (SCNI-SE01-P004). 

\end{ack}

\clearpage

\bibliography{references}
\bibliographystyle{apalike}

\newpage
\appendix
\section*{Appendices / Supplemental Materials}

\section{Additional Dataset Details}\label{sec:dataset_details}

    \subsection{Visual Summary}
\label{app:dataset_proportions}

\begin{figure}[htbp]
    \centering
    \includegraphics[width=\linewidth]{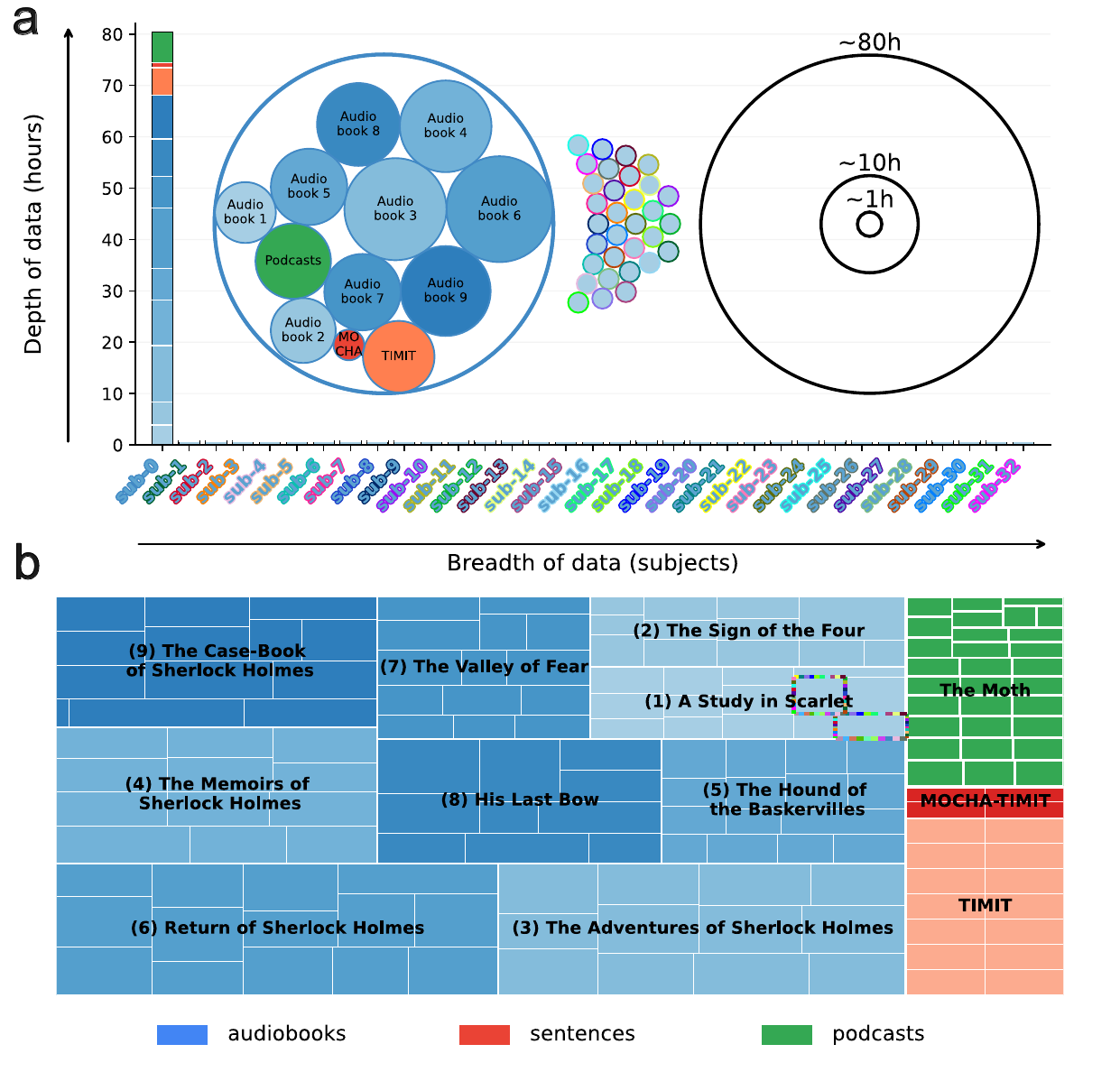}
    \caption{\textbf{Dataset proportions}.(\textbf{a}) Recording hours by subject and corpus. The stacked bar plot shows total duration per subject, with sub-0 comprising multiple linguistic materials and other subjects a subset of audiobook data. The bubble plot represents the same information (area proportional to hours); the larger sub-0 circle is subdivided into audiobooks, sentences, and podcasts, while smaller circles represent other subjects. Colours indicate data type, blue shades distinguish between audiobooks, and circle outlines distinguish between subjects. Black concentric circles provide a size reference (\textasciitilde 1 h, \textasciitilde 10 h, \textasciitilde 80 h). (\textbf{b}) Treemap of the full dataset, with rectangle area proportional to duration. Categories are subdivided into corpora and then into MEG recording sessions. The multicolour outline marks the only two audiobook sessions recorded from multiple subjects.}
    \label{fig:broad+deep}
\end{figure}

Figure~\ref{fig:broad+deep} provides two complementary views of the LibriBrain100 dataset composition. Panel (a) shows recording hours by subject and corpus; panel (b) shows a treemap of the full dataset with rectangle area proportional to duration. See Table~\ref{tab:data_splits} for a complementary perspective.

    \subsection{Comparison with Existing Datasets}
\label{app:existing_datasets}

Table~\ref{tab:datasets} lists existing MEG datasets for speech comprehension at the time of writing. LibriBrain100 compares favourably in total hours (104) and hours per subject (0.6--80), with the highest scores (by far, in the case of maximum hours per subject). It sits comfortably in the middle of the pack in terms of number of subjects, with 33 (range 1--96). The \texttt{pnpl} library contains data loaders for all of these datasets with support for reproducible data splits and for dataset downloading where possible, to support machine learning at scale. 

\begin{table}[h!]
\centering
\resizebox{\textwidth}{!}{%
\begin{tabular}{lllrrrrlll}
\toprule
\textbf{Name} & \textbf{Dataset} & \textbf{Language} & \textbf{Sensors} & \textbf{Total hrs} & \textbf{\# Subj} & \textbf{Hrs/Subj} & \textbf{Stimulus} & \textbf{Public} & \textbf{PNPL Dataloader} \\
\midrule
MOUS & \citet{schoffelen2019mous} & \cellcolor{orange!20}Dutch & 275 & 81 & \textbf{96} & 0.8 & Spoken sentences & \cellcolor{green!20}\href{https://data.ru.nl/collections/di/dccn/DSC_3011020.09_236}{Yes} & \cellcolor{green!20}Yes\\
MEG-MASC & \citet{gwilliams2023megmasc} & \cellcolor{red!10}English & 208 & 49 & 27 & 1.0 & MASC stories (synthetic voice) & \cellcolor{green!20}\href{https://doi.org/10.17605/OSF.IO/AG3KJ}{Yes} & \cellcolor{green!20}Yes\\
Armeni & \citet{armeni2022} & \cellcolor{red!10}English & 269 & 30 & 3 & 10.0 & LibriVox (Sherlock Book 3) & \cellcolor{green!20}\href{https://data.ru.nl/collections/di/dccn/DSC_3011085.05_995}{Yes} & \cellcolor{green!20}Yes \\
Le Petit Prince & \citet{dascoli2025natcomm} & \cellcolor{blue!10}French & \textbf{306} & 93 & 58 & 1.6 & Le Petit Prince & \cellcolor{green!20}\href{https://openneuro.org/datasets/ds007523/versions/1.0.1}{Yes} & \cellcolor{green!20}Yes \\
LibriBrain & \citet{ozdogan2025libribrain} & \cellcolor{red!10}English & \textbf{306} & 52 & 1 & 52.3 & LibriVox (Sherlock Book 1--7) & \cellcolor{green!20}\href{https://huggingface.co/datasets/pnpl/LibriBrain}{Yes} & \cellcolor{green!20}Yes \\
LibriBrain100 & (ours) & \cellcolor{red!10}English & \textbf{306} & \textbf{104} & 33 & 0.6--\textbf{80.0} & \makecell[tl]{LibriVox (Sherlock Book 1--9), \\TIMIT, MOCHA-TIMIT, \\The Moth (30 Podcasts)} & \cellcolor{green!20}\href{https://huggingface.co/datasets/pnpl/LibriBrain2}{Yes} & \cellcolor{green!20}Yes \\
\bottomrule
\\
\end{tabular}%
}
\caption{\textbf{Breadth and depth of speech comprehension MEG datasets}.}
\label{tab:datasets}
\end{table}

\section{Data Collection Methods}\label{sec:methods_extras}

    \subsection{Subjects}\label{sec:subjects}
MEG recordings were acquired from 33 volunteers (10 female, 23 male; age range 19–50 years, median = 28, interquartile range = 8). All participants reported normal hearing and normal or corrected-to-normal vision, and none had a history of neurological disorders. Sixteen participants were native speakers of English, while the remaining seventeen acquired English as a second language but were highly proficient and reported daily use for work or study. Prior to participation, all individuals provided informed consent for the use of anonymised data for research purposes. The study was approved by the University of Oxford Medical Sciences Interdivisional Research Ethics Committee (R90053/RE003).

    \subsection{Stimulus Materials}\label{sec:stimuli}
To prepare the audio stimuli for the MEG experiments, all audio files were converted to uncompressed WAV format and resampled to 48 kHz using SoX. The goal was to segment the audio into natural sentences or phrases, typically defined by pauses, and to align each segment with its corresponding textual content. This enabled the delivery of precise triggers to the MEG system during stimulus presentation, allowing continuous verification of temporal alignment between MEG recordings, audio signals, and linguistic annotations. Audio segmentation was carried out in two stages: an initial automated phase followed by manual refinement. The automated segmentation relied on voice activity detection (VAD), while the manual stage corrected segment boundaries in cases where VAD failed, for example by truncating low-intensity sounds such as utterance-initial voiceless plosives. In addition, manual refinement involved checking and reconciling consistency between data-driven audio segmentation and text-based boundaries (e.g., punctuation and sentence structure). VAD was implemented using custom scripts in Praat \citep{boersma2001}, identifying speech segments based on an intensity threshold of 59 dB and a minimum duration of 600 ms. The resulting TextGrid annotations were then populated with the corresponding text and carefully reviewed and corrected by hand. This process required over 200 hours of expert human effort, exceeding standard practices in the design and curation of comparable datasets, and was undertaken to ensure precise alignment across MEG recordings, audio signals, and linguistic annotations.

\subsubsection*{Audiobooks}%

Audio recordings for all nine books in the canon of Sherlock Holmes were sourced from LibriVox (\url{https://librivox.org/}). The books (including both novels and short stories) were presented in chronological order of publication. To minimise variance, we utilised recordings from the same reader (David Clarke) for books 1–8. As audio recordings were not available from this speaker for book 9, we selected a second reader with similar characteristics (Thomas A. Copeland; British English male). Each MEG recording session corresponded to a standalone chapter from the audiobooks. All texts are in the public domain and were obtained from Project Gutenberg (\url{https://www.gutenberg.org/}). Manual correction was used to fix typos and, where text and audio diverged, to align the text with the spoken audio. Ambiguous text, such as numbers, was normalised (e.g., “1492” rendered as “fourteen ninety-two” if spoken that way). Details for the audiobooks are provided in Table~\ref{tab:sherlock-audio}.

The Sherlock Holmes audiobooks comprise multiple detective fiction stories focused on solving mysteries through observation and logical deduction. Narratives typically involve identifying and interpreting clues, with cases ranging from concrete, crime-based investigations to stories that initially present more unusual or seemingly supernatural elements before being resolved through reasoning. The prose includes a mixture of narration and both direct and indirect dialogue. The thematics are relatively homogeneous (e.g., crime, investigation), but variation arises across books and chapters through differences in setting, narrative context, and secondary characters. The recordings are read by professional narrators, resulting in a controlled delivery: pauses tend to align with punctuation and sentence structure, diction is clear and consistent, and breathing and other non-linguistic artefacts are minimised.

\begin{table}[!htbp]
    \centering
    \begin{tabular}{|c|l|l|l|l|}
        \hline
        \textbf{Book} & \textbf{Name} & \textbf{Type} & \textbf{Sessions} & \textbf{Hours} \\
        \hline
        1 & \href{https://librivox.org/a-study-in-scarlet-version-6-by-sir-arthur-conan-doyle/}{A Study in Scarlet} (1888) & Novel & 14 & 04:37:34 \\
        2 & \href{https://librivox.org/the-sign-of-the-four-version-3-by-sir-arthur-conan-doyle/}{The Sign of the Four} (1890) & Novel & 12 & 04:27:31 \\
        3 & \href{https://librivox.org/the-adventures-of-sherlock-holmes-version-4-by-sir-arthur-conan-doyle/}{The Adventures of Sherlock Holmes} (1892) & Short Stories & 12 & 10:56:13 \\
        4 & \href{https://librivox.org/the-memoirs-of-sherlock-holmes-by-sir-arthur-conan-doyle-2/}{The Memoirs of Sherlock Holmes} (1893) & Short Stories & 12 & 08:53:17 \\
        5 & \href{https://librivox.org/the-hound-of-the-baskervilles-version-4-by-sir-arthur-conan-doyle/}{The Hound of the Baskervilles} (1901–1902) & Novel & 15 & 06:10:32 \\
        6 & \href{https://librivox.org/the-return-of-sherlock-holmes-by-sir-arthur-conan-doyle-2/}{Return of Sherlock Holmes} (1905) & Short Stories & 14 & 11:51:17 \\ 
        7 & \href{https://librivox.org/the-valley-of-fear-version-3-by-sir-arthur-conan-doyle/}{The Valley of Fear} (1914--1915) & Novel & 14 & 06:06:17 \\ 
        \hline
        8 & \href{https://librivox.org/his-last-bow-version-3-by-sir-arthur-conan-doyle/}{His Last Bow} (1917) & Short Stories & 10 & 07:10:29 \\ 
        9 & \href{https://librivox.org/the-case-book-of-sherlock-holmes-by-sir-arthur-conan-doyle/}{The Case-Book of Sherlock Holmes}* (1927) & Short Stories & 13 & 08:27:01 \\ 
        \hline
        \multicolumn{4}{c}{}&\multicolumn{1}{l}{\textbf{68:40:11}}\\ %
        \multicolumn{5}{l}{\footnotesize{*Read by Thomas A. Copeland}}\\ %
    \end{tabular}
    \caption{\textbf{Audiobooks available in LibriBrain100}. Books 1--7 appeared in the original LibriBrain release~\citep{ozdogan2025libribrain}; the rest are new here. Text in the name column link to the source audio on LibriVox.}
    \label{tab:sherlock-audio}
\end{table}

\subsubsection*{TIMIT}%

TIMIT is a corpus of read American English speech designed for acoustic–phonetic research and the development and evaluation of automatic speech recognition systems
\citep{garofolo1993darpa_timit,garofolo1993timit}. It contains recordings from 630 speakers across 8 major dialect regions of the United States, each producing 10 sentences, for a total of 6300 utterances ($\sim$5 hours of speech). All speakers were native speakers of American English and were screened to exclude clinically significant speech pathology. Speaker recruitment aimed to cover major dialect regions based on the region in which speakers lived during childhood. While the corpus achieves broad dialectal coverage, it is not perfectly balanced: some regions (e.g., New England, New York City, and “Army Brat”) are underrepresented due to practical constraints. The corpus is also sex-imbalanced (438 male, 192 female; approximately 70\%/30\%). 

The sentence prompts were designed to provide controlled phonetic coverage. Each speaker reads 2 shared SA (“dialect”) sentences, 5 SX sentences selected from a set of 450 phonetically compact sentences (each repeated across 7 speakers), and 3 SI sentences drawn from a set of 1890 phonetically diverse sentences (each produced by a single speaker), yielding 2342 distinct sentence texts in total. The SX and SI sentences were constructed to maximise coverage of phonetic contexts and allophonic variation, rather than semantic diversity. Speech was recorded at 16 kHz using 16-bit linear PCM encoding. In this dataset release, we provide MEG recordings for the full set of 6300 utterances. 

The TIMIT corpus consists of short, isolated sentences with no shared discourse context across utterances. The recordings are read speech, produced in a controlled setting, with clear articulation, limited disfluencies, and pauses aligned with sentence boundaries. At the same time, variability is introduced through the large number of speakers and dialect regions, providing diversity in pronunciation, accent, and voice characteristics. This combination of tightly controlled phonetic design and speaker variability makes TIMIT particularly well-suited for analyses focused on acoustic–phonetic representations.

\subsubsection*{MOCHA-TIMIT}%

MOCHA-TIMIT is a multichannel articulatory–acoustic corpus designed to support research on speech production and acoustic–articulatory modelling \citep{wrench1999mocha_timit,wrench2000mocha_timit}. The publicly distributed release contains data from two speakers: one male (msak0) and one female (fsew0), both Southern British English speakers. In this dataset release, we provide all 460 utterances from each speaker (920 utterances total). The sentence set was constructed to provide broad phonetic coverage while explicitly targeting connected-speech processes in English, such as assimilation and weak forms, and was designed to parallel the phonetic coverage of TIMIT while reflecting British English pronunciation. Although the MOCHA-TIMIT corpus was originally developed for studying speech production, the tight control over articulation and phonetic content also makes it well-suited for perception studies, as it provides highly structured and phonetically balanced stimuli with well-characterised acoustic realisations.

Each utterance is a short, read sentence produced in a controlled recording environment. As in TIMIT, there is no extended narrative context across sentences; instead, the corpus emphasises phonetic and articulatory diversity within isolated utterances. The recordings are carefully produced, with clear diction, minimal disfluencies, and pauses aligned with sentence boundaries. Compared to TIMIT, however, the speech more systematically reflects connected-speech phenomena, resulting in more natural coarticulation patterns despite the controlled setting. For example, assimilation processes may occur across word boundaries (e.g., “good boy” realised with a more bilabial /d/ influenced by the following /b/), and weak forms are frequently used for function words (e.g., ``and'' $\rightarrow$ /\textipa{@n}/, ``to'' $\rightarrow$ /\textipa{t@}/, ``of'' $\rightarrow$ /\textipa{@v}/). Vowel reduction and consonant lenition are also present in unstressed positions, and segmental realisations are influenced by surrounding phonetic context, yielding more continuous and context-dependent articulatory patterns.

\subsubsection*{Podcasts}%

We used 30 stories (6 hours) from The Moth podcast, drawn from a larger set of 77 stories ($\sim$15 hours) previously used for semantic decoding from fMRI data ~\citep{tang2023}. This subset was selected using a data-driven criterion aimed at maximising semantic coverage. Specifically, we estimated the spread of each story in semantic space using sentence embedding models, which map sentences to dense vector representations that capture their semantic content. For each story, we computed the centroid of its sentence embeddings and quantified dispersion as the mean cosine distance of individual sentence vectors to this centroid in the high-dimensional embedding space. Stories with greater dispersion (i.e., covering a wider range of semantic content, as opposed to being concentrated in a narrow region of the embedding space) were preferentially selected. Details for the selected stories are provided in Table~\ref{tab:moth-stories}.

The Moth podcast stories provide an effective set of naturalistic stimuli for neural decoding, in particular for semantic representations, because they combine ecological validity with high semantic diversity and coherent narrative structure. Each stimulus consists of a single speaker telling an autobiographical narrative, ensuring continuity in voice, perspective, and discourse structure. Crucially, the corpus spans a wide range of topics and contexts, from historically and geographically specific settings (e.g., Zimbabwe during independence, wartime Baghdad, travel in the USSR, scientific expeditions) to personal and emotional experiences (e.g., bereavement, parenthood, moral dilemmas), as well as humorous, reflective, and self-development narratives. This breadth yields rich variation in entities, events, and abstract themes, providing dense coverage of semantic space. At the same time, individual stories are internally coherent and contextually well-defined, allowing models to track how meaning unfolds over extended timescales.

These narratives are delivered as spontaneous speech rather than read text, and therefore include natural disfluencies such as interjections (e.g., “like,” “you know,” “I mean”), repetitions (e.g., “and and then…,” “I was I was…”), and repairs (e.g., “on Tuesday—uh, Wednesday,” “her brother—no, her cousin”), which are largely absent in scripted speech. Temporal dynamics also differ from read speech: pauses can reflect on-the-fly planning, while other segments are produced in dense, continuous stretches, a characteristic feature of spontaneous narration. In addition, prosodic variation is more pronounced, as speakers may raise their voice or modulate intonation for emphasis or comedic effect. Recordings also include audience responses such as laughter and applause, which introduce acoustic variability but can carry contextual and pragmatic information tied to the narrative. Together, these properties enhance ecological validity and introduce variability that better reflects real-world language processing. The combination of rich thematic diversity, extended context, and naturalistic delivery, as leveraged in prior work ~\citep[e.g.,][]{tang2023}, makes Moth stories particularly well-suited for training and evaluating neural decoders that map brain activity to continuous, context-sensitive semantic representations.

{\scriptsize
\begin{longtable}{@{}r p{0.32\linewidth} p{0.20\linewidth} r p{0.08\linewidth}@{}}
\caption{\textbf{Podcasts available in LibriBrain100}. Story titles are hyperlinked to the source audio and transcript at \url{https://themoth.org/stories/}. Validation and test set stories are highlighted \sethlcolor{yellow!50}\hl{yellow} and \sethlcolor{red!50}\hl{red}, respectively.  Durations are given in minutes and seconds for the audio files used in our stimulus set. }
\label{tab:moth-stories}\\
\hline
\# & \textbf{Story} & \textbf{Author} & \textbf{Duration}\\
\hline
\endfirsthead
\hline
\# & \textbf{Story} & \textbf{Author} & \textbf{Duration}\\
\hline
\endhead
\hline
\endfoot
\hline

\multicolumn{3}{r}{\textbf{}} & \textbf{6:00:59} \\

\endlastfoot

1  & \href{https://themoth.org/stories/alternate-ithaca-tom}{Alternate Ithaca Tom} & Tom Weiser & 11:47 \\ %
2 & \href{https://themoth.org/stories/thumbs-up}{Thumbs Up!} & Nathan Englander & 14:09 \\ %
3 & \href{https://themoth.org/stories/goldie-the-goldfish}{Goldie, the Goldfish} & Becca Stevens & 10:54 \\ %
4  & \href{https://themoth.org/stories/under-the-influence}{Under the Influence} & Jeffery Rudell & 10:28 \\ %
5 & \href{https://themoth.org/stories/treasure-island}{Treasure Island} & Boots Riley & 13:30 \\ %
6 & \href{https://themoth.org/stories/that-thing-on-my-arm}{That Thing on My Arm} & Padma Lakshmi & 14:49 \\ %
7 & \href{https://themoth.org/stories/breaking-up-in-the-age-of-google}{Breaking Up in the Age of Google} & Jessi Klein & 17:43 \\ %
8 & \href{https://themoth.org/stories/the-triangle-shirtwaist-connection}{The Triangle Shirtwaist Connection} & Michelle Fecteau & 7:06 \\ %
9 & \href{https://themoth.org/stories/stage-fright}{Stage Fright} & Suzanne Vega & 10:07 \\ %
10 & \href{https://themoth.org/stories/not-on-the-usual-tour}{Not on the Usual Tour} & Jon Lovett & 8:31 \\ %
11 & \href{https://themoth.org/stories/life-reimagined}{Life Reimagined} & Raymond Christian & 11:15 \\ %
12 & \href{https://themoth.org/stories/only-one-way-to-find-out}{Only One Way To Find Out} & Sarah Gray & 13:04 \\ %
13 & \href{https://themoth.org/stories/the-postman-always-calls}{The Postman Always Calls} & Elizabeth Browning & 15:29 \\ %
14 & \href{https://themoth.org/stories/blue-hope}{Blue Hope} & Sylvia Earle & 14:00 \\ %
15 & \href{https://themoth.org/stories/the-curse}{The Curse} & Dame Wilburn & 13:56 \\ %
16 & \href{https://themoth.org/stories/clifton-truman-daniel}{Beneath the Mushroom Cloud} & Clifton Truman Daniel & 11:46 \\ %
17 & \href{https://themoth.org/stories/going-the-liberty-way}{Going the Liberty Way} & Kevin Roose & 13:28 \\ %
18 & \href{https://themoth.org/stories/caution-eating}{Caution: Eating} & Evan Kleiman & 9:40 \\ %
19 & \href{https://themoth.org/stories/birth-of-a-nation}{Birth of a Nation} & Petina Gappah & 9:09 \\ %
20 & \href{https://themoth.org/stories/life-and-death-on-the-oregon-trail}{Life and Death on the Oregon Trail} & Micaela Blei & 12:24 \\ %
21 & \href{https://themoth.org/stories/fire-test-for-love}{Fire Test For Love} & Lauren Slater & 10:58 \\ %
22 & \href{https://themoth.org/stories/leaving-baghdad}{Leaving Baghdad} & Abbas Mousa & 11:15 \\ %
23 & \href{https://themoth.org/stories/coming-of-age-on-death-row}{Coming of Age on Death Row} & Gautam Narula & 11:58 \\ %
24  & \href{https://themoth.org/stories/how-to-draw-a-nekkid-man-storyslam-version}{How To Draw A Nekkid Man} & Tricia Rose Burt & 12:08 \\ %
25 & \href{https://themoth.org/stories/mayor-of-freaks}{The Mayor of the Freaks} & David Crabb & 16:11 \\ %
26 & \href{https://themoth.org/stories/swimming-with-astronauts}{Swimming with Astronauts} & Michael J.\ Massimino & 13:11 \\ %
27 & \href{https://themoth.org/stories/wild-womxn-and-dancing-queens}{Wild Womxn and Dancing Queens} & Lex Jade & 6:40 \\ %
28 & \href{https://themoth.org/stories/vixen-and-the-ussr}{Vixen and the USSR} & Sue Steinacher & 13:24 \\ %
\rowcolor{yellow!50} 29 & \href{https://music.apple.com/pa/album/the-best-of-the-moth-vol-7/488876794?l=en}{From Boyhood to Fatherhood} & Jonathan Ames & 11:57 \\ %
\rowcolor{red!50} 30 & \href{https://themoth.org/stories/where-theres-smoke}{Where There's Smoke} & Jenifer Hixson & 10:02 \\ %
\end{longtable}
}

    \subsection{Experimental Design \& Procedure}\label{sec:protocol}

Each recording session began with visually presented instructions projected onto a translucent whiteboard using a DLP LED projector (ProPixx, VPixx Technologies Inc., Saint-Bruno, Canada). Participants were seated inside the MEG scanner and initiated the experiment via button press. Auditory stimuli were delivered binaurally through non-metallic air tube earphones (Aero Technologies) at approximately 70 dB SPL, with minor adjustments to bass and treble based on participant preference. Stimulus delivery was controlled using the PsychoPy toolbox \citep{peirce2007psychopy}, and the stimulus computer synchronized with the MEG acquisition system via a parallel port to send precise event triggers marking stimulus onset with millisecond temporal accuracy.

For the multiple subjects cohort (sub-01 to sub-32), all participants listened to Chapters 11 and 12 of A Study in Scarlet. Both chapters were presented within a single recording session lasting approximately one hour, with a short break between chapters. To assess attention and comprehension, participants answered five questions at the end of each chapter. Each question was presented in a four-alternative multiple-choice format with distractors (e.g., “What does Jefferson Hope take from Lucy Ferrier after her death?” with answer options “A: A locket, B: A necklace, C: A wedding ring, D: A bracelet”). Responses were recorded via button press using a MEG-compatible ResponsePixx Dual Handheld system (VPixx Technologies Inc., Saint-Bruno, Canada).

Subject 0 participated in multiple recording sessions and was exposed to a broader set of linguistic materials, including all books in the Sherlock Holmes canon, as well as additional speech corpora (e.g., TIMIT, MOCHA-TIMIT, and The Moth podcasts). Comprehension assessment varied by stimulus type. For audiobook stimuli, a single comprehension question was presented at the end of each chapter in a two alternative options format (e.g., “Where is the body of the murder victim found?” with answer options “A: in the bedroom”, “B: in the garden”). For TIMIT and MOCHA-TIMIT, questions were presented immediately after selected sentences (approximately 20 questions per session; ~5\% of trials for TIMIT, ~10\% for MOCHA-TIMIT). These questions were presented in a four alternative multiple options format (e.g., “\_\_\_\_\_\_ is a pop singer.” with answer options “A: Michael Jackson, B: Tina Turner, C: Elton John, D: Madonna”; “Clear \_\_\_\_\_\_ is appreciated.” with answer options “A: grammar, B: diction, C: articulation, D: pronunciation”) and typically targeted a key word (often a noun), with distractors that were either acoustically or semantically similar. For the podcast corpus, five comprehension questions were presented at the end of each episode, probing key details, event sequences, and overall meaning (e.g., “How did thinking about his alternate life affect him?” with answer options “A: It made him confident, B: It made him confused and unhappy, C: It motivated him to study, D: It made him excited”). For subject 0, each recording session lasted approximately three hours and typically included either 3–5 audiobook chapters, 1000–1200 sentences, or 8–10 podcast episodes. Short breaks were allowed between recording blocks. Recording sessions were spaced at least one day apart, with no more than two months between sessions, depending on participant and experimenter availability.

    \subsection{Data Acquisition}\label{sec:acquisition}
Prior to data acquisition, each participant’s head shape was digitised using a Polhemus Fastrak 3D digitiser (Polhemus, Vermont, USA). This procedure included the localisation of fiducial landmarks (nasion, left and right pre-auricular points), along with approximately 300 additional points sampled across the scalp, forehead, and nose. Five Head Position Indicator (HPI) coils were positioned on the mastoid and forehead to enable continuous monitoring of head position during MEG recording via electromagnetic induction. MEG recordings were acquired using a MEGIN Triux™ Neo system (York Instruments Ltd., Heslington, UK), consisting of 102 magnetometers and 204 orthogonal planar gradiometers. The system was installed in a magnetically shielded room to reduce environmental interference. Prior to entering the recording environment, participants were screened for metallic objects and other potential sources of electromagnetic noise. During acquisition, participants were seated with their head positioned close to the dewar. Participants were instructed to minimise head, body, and limb movements throughout the recording. Data were sampled at 1000 Hz with an online band-pass filter of 0.01–330 Hz. Ocular activity was monitored using bipolar electrooculogram (EOG) electrodes, with one pair placed at the outer canthi (horizontal EOG) and another above and below the left eye (vertical EOG). Cardiac signals were recorded via bipolar electrocardiogram (ECG) electrodes positioned on the clavicle and hip. Articulatory muscle activity was continuously tracked using electromyography (EMG), with electrodes placed below the cheekbone (jaw movement), below the lower lip (lip movement), and beneath the chin to capture potential laryngeal activity.

    \subsection{Minimal Preprocessing Pipeline}\label{sec:preprocessing}

The MEG data were minimally preprocessed to remove measurement artefacts while preserving flexibility for additional preprocessing steps for downstream analyses. Head position was estimated using continuous recordings from the Head Position Indicator (HPI) coils throughout each recording session. This information was used to correct for head movements, and data from all participants were realigned to a common reference head position. This procedure was applied consistently across subjects and is expected to reduce variability due to head position differences and within-session head movement. Noisy channels were identified, excluded, and subsequently reconstructed via interpolation from neighbouring sensors. Environmental noise was attenuated using Maxwell filtering. Specifically, a temporally non-extended Signal Space Separation (SSS) algorithm \citep{taulu2006spatiotemporal} was applied to suppress sources originating outside the head. In contrast, no explicit steps were taken to remove physiological artefacts (e.g., eye blinks, cardiac activity, or muscle contractions), allowing users to apply additional artefact correction procedures as appropriate for their specific downstream analyses. Power line noise at 50 Hz and its 100 Hz harmonic was removed using notch filters. The data were then band-pass filtered between 0.1 and 125 Hz using zero-phase, two-pass Butterworth filters to reduce slow drifts and prevent aliasing effects prior to downsampling. Finally, the data were downsampled to 250 Hz, yielding recordings with 306 channels and a temporal resolution of 4 ms per sample.

    \subsection{Event Files/Labels}\label{sec:labels}

Annotations were generated for each session, providing onset times and durations (in seconds) for multiple event types, including silence (i.e., non-speech segments, which may include breathing and occasional laughter or applause), word-level units (e.g., A, Study, in, Scarlet), and phoneme-level units (e.g., ah, s, t, ah, d, iy, ih, n, s, k, aa, r, l, ah, t). Additional annotation fields include phoneme position within words, encoded using conventions from Kaldi \citep{povey2011kaldi}: B (beginning), I (inside), E (end), and S (singleton). Each MEG recording session (available as a FIF file) is accompanied by a corresponding event file in TSV format, which can be used to define labels for supervised learning tasks. These event files were derived from the transcripts and their corresponding audio segments.

To obtain precise temporal alignment between text and audio, forced alignment was performed using Gentle \citep{gentle}, a Kaldi-based aligner that employs Gaussian Mixture Model–Hidden Markov Model (GMM-HMM) architectures to combine acoustic and language models and determine the most likely alignment path between audio and transcript. Because the audio had already been segmented into short utterances using VAD, as described above, the forced alignment step was simplified. Instead of processing entire chapters, the aligner was applied to short, pre-segmented audio–transcript pairs, improving both robustness and alignment accuracy.

Gentle produces phoneme-level annotations in the ARPABET format, consisting of a set of 39 phoneme categories. ARPABET provides a discrete, standardized phonemic representation that captures the more abstract, relatively invariant characteristics of speech sounds, but does not capture finer phonetic detail such as allophonic variation or diphthong structure. The same forced-alignment procedure was applied consistently across all linguistic materials (audiobooks, TIMIT, MOCHA-TIMIT, and podcasts) to ensure internal consistency within the dataset. Although alternative phonetic annotations (e.g., IPA transcriptions) are available in the standard releases of the TIMIT and MOCHA-TIMIT corpora, they were not used to generate event files in this dataset. Instead, a unified ARPABET representation was adopted to maintain consistency across heterogeneous linguistic materials.

The forced aligner occasionally produced suboptimal outputs, particularly in challenging cases. Failures were most frequent for proper names and other out-of-vocabulary (OOV) items, quotations in different languages, and segments with atypical prosody (e.g., voice imitation or exaggerated emphasis). Additional difficulties arose in more spontaneous speech (e.g., podcasts), including repetitions, repairs, and unclear or reduced pronunciations. These cases were identified and corrected through manual inspection. Audio segments were reviewed by human experts using Praat \citep{boersma2001}, with spectrograms and waveforms examined to refine word and phoneme boundaries and recover missed lexical items. As a result of this intervention, all word-level annotations were recovered. Following manual correction, the proportion of OOV annotations was substantially reduced for audiobooks and podcasts. Overall, this process resulted in high-quality, densely annotated data suitable for both word-level and phoneme-level analyses.

    \subsection{Standard Data Splits}\label{sec:splits}
\paragraph{Audiobooks.} For Subject 0, we reuse the same validation and test splits defined in LibriBrain~\citep{ozdogan2025libribrain}. Here, Sherlock book 1, session 11 is used for validation and session 12 is used for test. As subjects 1-32 have no official training data, when training on these subjects, we use the first half of book 1, session 11 for training, the second half for validation, and keep session 12 independent for testing. When training multi-subject data jointly with Subject 0, we also use only the second half of Subject 0's session 11 for validation, adding the first half to the training set.

\paragraph{TIMIT.} We use the official TIMIT core test speaker set of 24 speakers for our test set and 50 speakers from the dev split \citep{garofolo1993timit}. We exclude all \texttt{SA} labelled sentence IDs from testing and validation as these are repeated by all speakers (including those in training). We also exclude any speakers from training who have sentences that overlap with the speakers in the test set. See Table~\ref{tab:timit-speakers} for a complete list of the speakers we use in validation and test.

\begin{table}[htbp]
  \centering
  \caption{\textbf{TIMIT speaker IDs used for the test and validation splits}.}
  \label{tab:timit-speakers}
  \begin{tabular}{@{}lllll@{}}
    \toprule
    \multicolumn{5}{c}{\textbf{Test set} (24 speakers)} \\
    \midrule
    \texttt{mdab0} & \texttt{mwbt0} & \texttt{felc0} & \texttt{mtas1} & \texttt{mwew0} \\
    \texttt{fpas0} & \texttt{mjmp0} & \texttt{mlnt0} & \texttt{fpkt0} & \texttt{mlll0} \\
    \texttt{mtls0} & \texttt{fjlm0} & \texttt{mbpm0} & \texttt{mklt0} & \texttt{fnlp0} \\
    \texttt{mcmj0} & \texttt{mjdh0} & \texttt{fmgd0} & \texttt{mgrt0} & \texttt{mnjm0} \\
    \texttt{fdhc0} & \texttt{mjln0} & \texttt{mpam0} & \texttt{fmld0} &              \\
    \midrule
    \multicolumn{5}{c}{\textbf{Validation set} (50 speakers, from dev split)} \\
    \midrule
    \texttt{faks0} & \texttt{fdac1} & \texttt{fjem0} & \texttt{mgwt0} & \texttt{mjar0} \\
    \texttt{mmdb1} & \texttt{mmdm2} & \texttt{mpdf0} & \texttt{fcmh0} & \texttt{fkms0} \\
    \texttt{mbdg0} & \texttt{mbwm0} & \texttt{mcsh0} & \texttt{fadg0} & \texttt{fdms0} \\
    \texttt{fedw0} & \texttt{mgjf0} & \texttt{mglb0} & \texttt{mrtk0} & \texttt{mtaa0} \\
    \texttt{mtdt0} & \texttt{mthc0} & \texttt{mwjg0} & \texttt{fnmr0} & \texttt{frew0} \\
    \texttt{fsem0} & \texttt{mbns0} & \texttt{mmjr0} & \texttt{mdls0} & \texttt{mdlf0} \\
    \texttt{mdvc0} & \texttt{mers0} & \texttt{fmah0} & \texttt{fdrw0} & \texttt{mrcs0} \\
    \texttt{mrjm4} & \texttt{fcal1} & \texttt{mmwh0} & \texttt{fjsj0} & \texttt{majc0} \\
    \texttt{mjsw0} & \texttt{mreb0} & \texttt{fgjd0} & \texttt{fjmg0} & \texttt{mroa0} \\
    \texttt{mteb0} & \texttt{mjfc0} & \texttt{mrjr0} & \texttt{fmml0} & \texttt{mrws1} \\
    \bottomrule
  \end{tabular}
\end{table}

\paragraph{MOCHA-TIMIT.} As there is no standard split for MOCHA-TIMIT, we define our own split, being careful to avoid any sentence stimulus overlap while maintaining an independent test session. We note that MOCHA-TIMIT contains four different sets: A, B, C, and D where sentences are repeated between A and D, and between B and C in shuffled orders. We use all of set A and D for training, then split B and C into validation and test as follows. We take the first half of the sentences in set B as the validation set. We then use the sentences in set C that are not in the first half of set B as test data, noting that these test sentences will not be consecutive within set C. 
In the Hugging Face repository, sets A, B, C, D correspond to sessions 1, 2, 3, 4. 

\paragraph{Podcasts.} Following \citet{tang2023}, two stories were designated to be held-out from training. For reproducibility we thus included \textit{From Boyhood to Fatherhood} in our validation set and \textit{Where There's Smoke} in our test set. All other stories should be included in training.

\section{Additional Stimulus Details}\label{sec:stim_extras}

    \subsection{Linguistic Variability}\label{sec:linguistic}

One of the central objectives of this dataset is to enable rigorous tests of generalization in neural decoding of speech perception. In the context of MEG, this is particularly important because models are exposed to substantial variability in natural speech, spanning differences in speakers, phonetic realisations, and semantic content. Rather than aiming to isolate invariant representations, the goal is to evaluate how well models can achieve robust performance in the presence of such variability, potentially leveraging both variable and more stable features of the signal. To this end, the dataset was designed to span multiple dimensions of variability that are known to shape speech processing, including speaker-dependent acoustic properties, phonetic composition, and higher-level semantic structure. By systematically varying these factors across complementary corpora, we can assess the extent to which decoding models remain resilient across changes in linguistic context and input statistics. This is especially relevant for tasks such as phoneme and word classification, where strong performance should be maintained despite differences in speakers, phonetic distributions, and semantic environments. In addition, the combination of variability and scale allows us to examine how exposure to diverse speech inputs influences learning dynamics, for instance whether it supports faster convergence toward more robust representations that better reflect the range of conditions encountered in natural speech perception.

\begin{figure}
    \centering
    \includegraphics[width=0.8\linewidth]{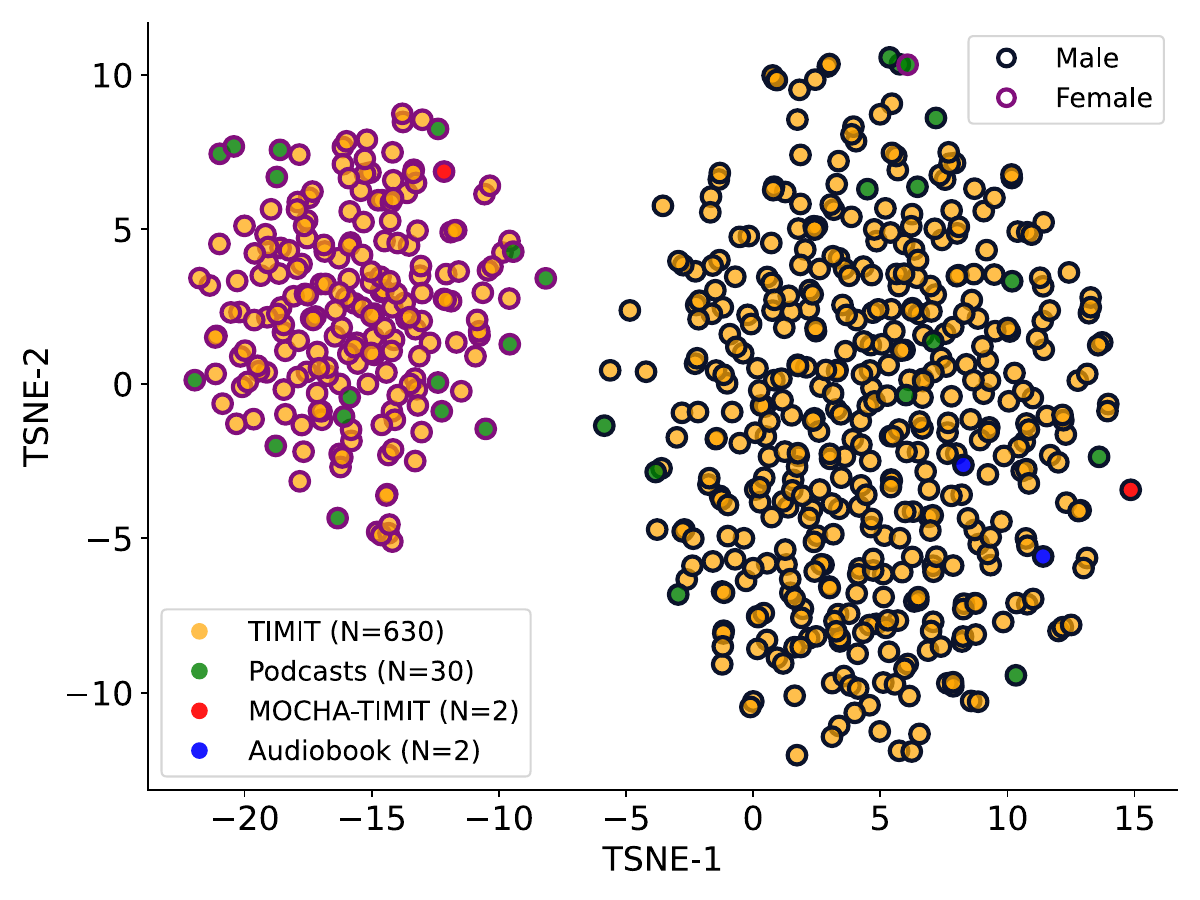}
    \caption{\textbf{Speaker embeddings}. Two-dimensional t-SNE projection of ECAPA-TDNN speaker embeddings, which capture acoustic properties such as timbre, pitch, and vocal tract characteristics. Each point represents a single speaker, obtained by averaging embeddings across multiple utterances. Marker fill colour indicates the speech corpus (TIMIT, Podcasts, MOCHA-TIMIT, Audiobook), while marker edge colour denotes speaker sex (male, female).}
    \label{fig:voices}
\end{figure}

At the acoustic level, one of the main goals of the dataset is to introduce substantial variability in speaker characteristics, thereby enabling a more realistic assessment of robustness in speech perception. To quantify this variability, we performed a speaker embedding analysis using a pretrained ECAPA-TDNN model \citep{desplanques2020ecapa}. These embeddings capture voice-related acoustic properties such as pitch, timbre, and vocal tract characteristics, providing a compact representation of speaker differences. When projected into a low-dimensional space (Figure~\ref{fig:voices}), the embeddings reveal a clear separation between male and female speakers, indicating that they capture meaningful acoustic structure. This separation is expected, as one of the most salient differences in speech arises from fundamental frequency (F0) and vocal tract size: on average, male speakers have larger vocal tracts and produce lower-pitched sounds, whereas female speakers tend to have smaller vocal tracts and higher-pitched voices. Beyond this dominant axis, the embeddings also reflect finer-grained differences between individual speakers, consistent with their intended role in capturing speaker identity. Notably, TIMIT spans a broader region of this space compared to the other corpora, indicating wider coverage of speaker variability, in part due to its large number of speakers. Combining all corpora further expands this coverage, resulting in a richer sampling of pitch- and timbre-related dimensions. This is particularly relevant for MEG neural decoding, as such acoustic features are represented in auditory cortex and contribute to the variability of the neural signal. By exposing models to this range of speaker-dependent variation, the dataset allows us to assess how well decoding performance is maintained across acoustic differences, and establishes a foundation for variability not only at the acoustic level but also in the phonetic realisations of speech. \\

\begin{figure}
    \centering
    \includegraphics[width=1.0\linewidth]{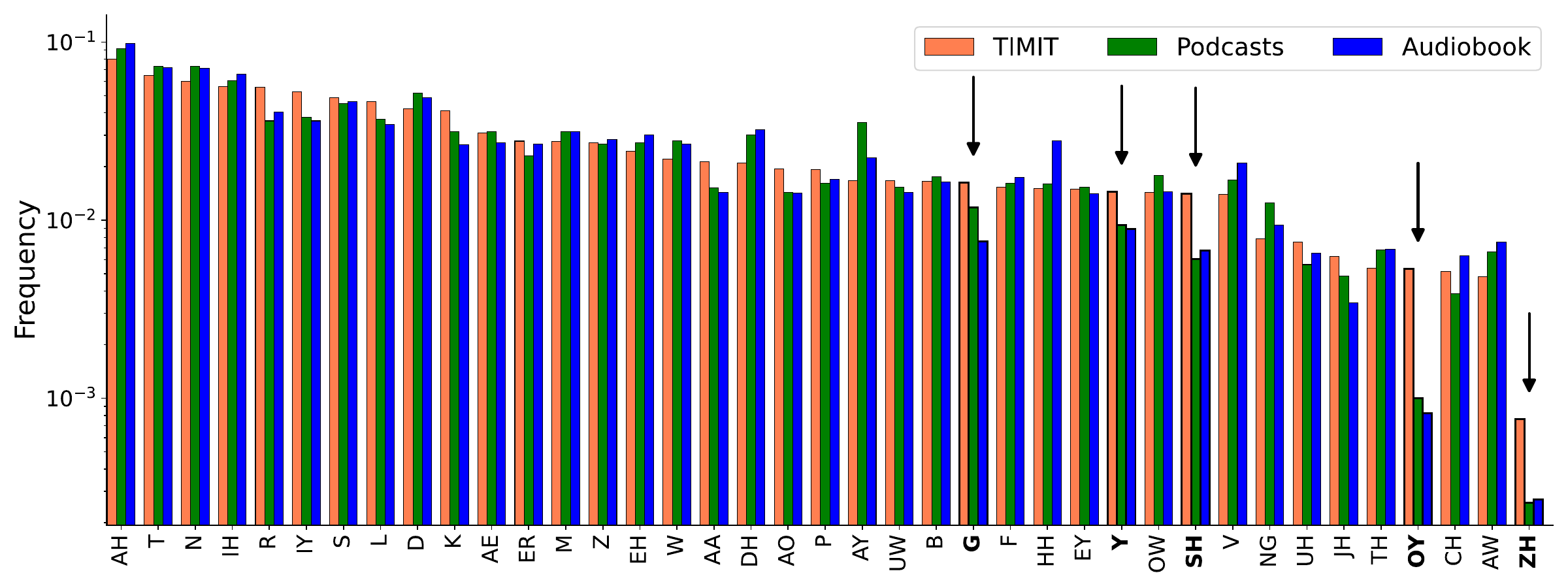}
    \caption{\textbf{Phoneme frequency distribution per corpus}. Bar plots show normalised phoneme frequencies for each corpus, sorted by decreasing frequency in TIMIT. Each phoneme category on the x-axis contains three adjacent bars corresponding to TIMIT (orange), Podcasts (green), and Audiobook (blue). The y-axis is displayed on a logarithmic scale to enhance visibility of low-frequency phonemes. Selected low-frequency phonemes (G, Y, SH, OY, ZH) are highlighted with downward-pointing arrows and thicker bar outlines to facilitate comparison across corpora.}
    \label{fig:frqdist}
\end{figure}

\begin{figure}[!htbp]
    \centering
    \includegraphics[width=0.8\linewidth]{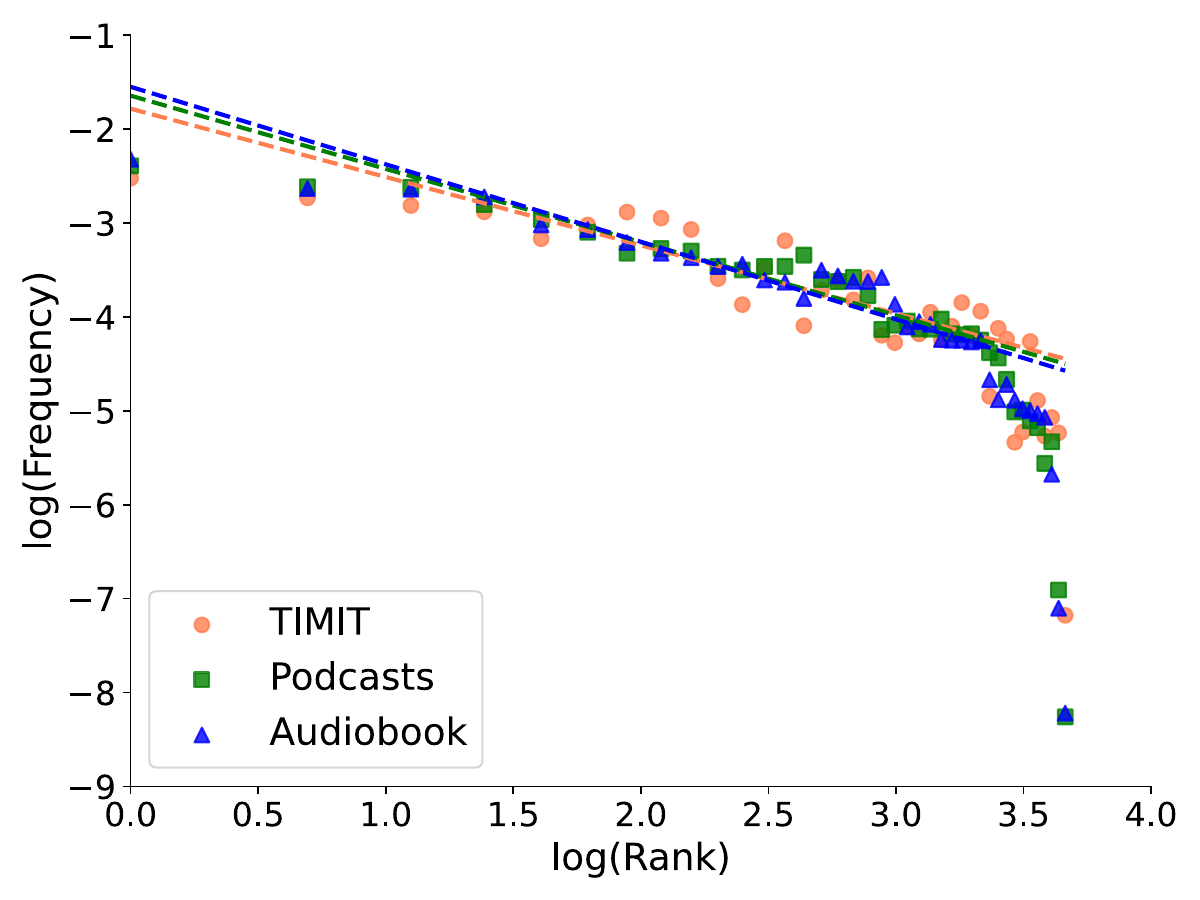}
    \caption{\textbf{Phoneme rank-frequency distribution per corpus}. Each point represents a phoneme, with its position determined by its rank (x-axis) and normalised frequency (y-axis), both shown in logarithmic scale. Phonemes are ranked in decreasing order of frequency based on TIMIT. Dashed lines indicate linear fits computed over ranks. Different marker shapes (circles, squares, triangles) and different colors (orange, green, blue) distinguish the three corpora.}
    \label{fig:zipf}
\end{figure}

At the phonetic level, this variability extends to the distribution and realisation of speech sounds, providing a complementary axis along which robustness can be evaluated. TIMIT, podcasts, and audiobooks offer distinct phonetic properties, with TIMIT in particular engineered to achieve a more balanced coverage of phonemes, including relatively rare sounds. This is accomplished by repeatedly including sentences that contain less frequent phonemes, thereby increasing their occurrence while preserving the structure of natural speech. As shown in Fig.~\ref{fig:frqdist}, using approximately 200,000 phoneme tokens per corpus, noticeable differences emerge in the frequency of rare phonemes such as G, Y, SH, OY, and ZH compared to the other corpora. This pattern is also reflected in the rank–frequency analysis (Fig.~\ref{fig:zipf}), where the fitted slopes indicate a flatter distribution for TIMIT ($-0.7271$) relative to Podcasts ($-0.7796$) and Audiobook ($-0.8254$), implying reduced dominance of high-frequency phonemes and increased representation of low-frequency ones. Crucially, the large number of speakers in TIMIT further amplifies phonetic variability by introducing a wide range of accents, speaking styles, and speaker-specific articulations. As a result, individual phonemes are realised through a richer set of allophonic variants, and co-articulatory patterns vary more extensively across contexts. This combination of balanced phoneme frequencies and diverse phonetic realisations leads to a substantially broader coverage of acoustic–phonetic patterns. From the perspective of MEG-based neural decoding, this increased variability provides a stronger test of whether models can maintain performance across both shifts in phoneme statistics and differences in their realisation, and whether increased diversity and scale facilitate the learning of representations that remain effective across heterogeneous linguistic inputs. \\

\begin{figure}[!htbp]
    \centering
    \includegraphics[width=0.9\linewidth]{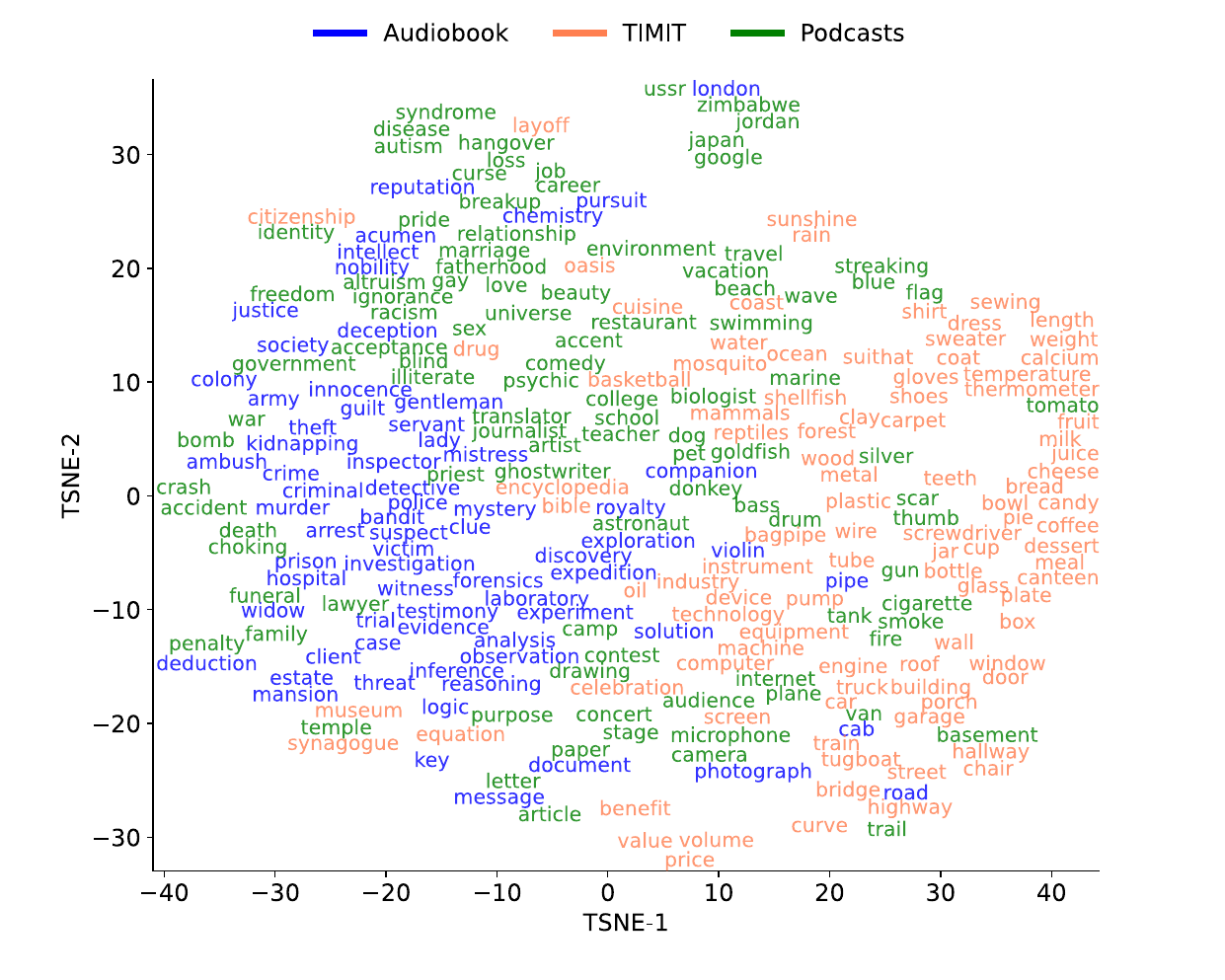}
    \caption{\textbf{Keyword embeddings per corpus}. Two-dimensional t-SNE projection of keyword embeddings from the Audiobook (blue), TIMIT (orange), and Podcasts (green) corpus . Each point corresponds to a keyword, and proximity between points reflects similarity in the embedding space, capturing lexical-semantic relationships between words.}
    \label{fig:keywords}
\end{figure}

Finally, at the semantic level, the three corpora differ in the richness and diversity of their lexical content, providing a higher-level dimension of variability. The audiobook corpus is relatively constrained, as it centres on Sherlock Holmes narratives, leading to redundancy in settings, characters, and thematic structure. TIMIT, by contrast, spans a broad range of topics but consists of short, largely unrelated sentences, limiting coherence across samples. The podcast corpus provides the greatest semantic diversity, comprising multiple extended narratives on distinct topics. These differences are illustrated in Fig.~\ref{fig:keywords}, where a pretrained Word2Vec model \citep{mikolov2013distributed} was used to extract word embeddings for representative keywords. The embeddings were normalised and projected into two dimensions using cosine distance. In semantic space, keywords show a compact cluster for the audiobook corpus and more dispersed distributions for the TIMIT and, especially, the podcasts corpus. More generally, the corpora occupy partially distinct regions of the semantic embedding space, reflecting differences in lexical content. Taken together, they provide broader semantic coverage than any individual corpus alone. This diversity allows us to examine how decoding models perform across variations in meaning and discourse context, and whether exposure to semantically richer and more varied inputs supports more robust performance. In combination with the other sources of variability, this also enables us to investigate how scale and diversity jointly influence the stability of learned neural representations in naturalistic speech perception.

\section{Additional MEG Details}\label{sec:meg_extras}

    \subsection{Neural Variability}\label{sec:neural_variability}
\raggedbottom

A challenge in decoding speech perception from MEG lies in the variability of the neural signal itself. Unlike controlled experimental paradigms, naturalistic listening introduces fluctuations that arise not only from the stimulus, but also from the listener. In the present dataset, this variability is intrinsic to the design: recordings span multiple participants and multiple sessions per participant, each associated with differences in attention, engagement, fatigue, and individual neurophysiology. Participants vary in language background (native vs. non-native), in their level of sustained attention to the narrative, and in cognitive states such as drowsiness or mind wandering. Even within the same individual, these factors can fluctuate across recording sessions, especially when sessions are conducted consecutively, leading to changes in alertness and engagement. In addition, inter-individual differences in brain anatomy, head shape, and age-related factors influence the spatial configuration and magnitude of the measured MEG signals. Together, these sources of variability shape the observed neural responses and constitute a fundamental challenge for any model aiming to extract robust neural representations of speech.

To characterise this variability, we examined three complementary dimensions of the MEG signal — amplitude, phase, and power — contrasting speech and non-speech segments. These measures capture distinct aspects of auditory cortical processing. The root mean square (RMS) of the evoked response reflects the strength of stimulus-locked activity, which is typically enhanced over bilateral temporal sensors during speech perception. Inter-trial coherence (ITC) quantifies the consistency of phase alignment across trials, and is known to increase at low frequencies when neural activity entrains to the temporal structure of speech. Finally, power spectral density (PSD) provides a frequency-resolved measure of oscillatory activity, with speech perception commonly associated with increased power in the delta–theta frequency band (\textasciitilde 2–7 Hz), corresponding to prosodic and syllabic rhythms \citep{peelle2012neural}.

\begin{figure}[H]
    \centering
    \includegraphics[width=0.98\linewidth]{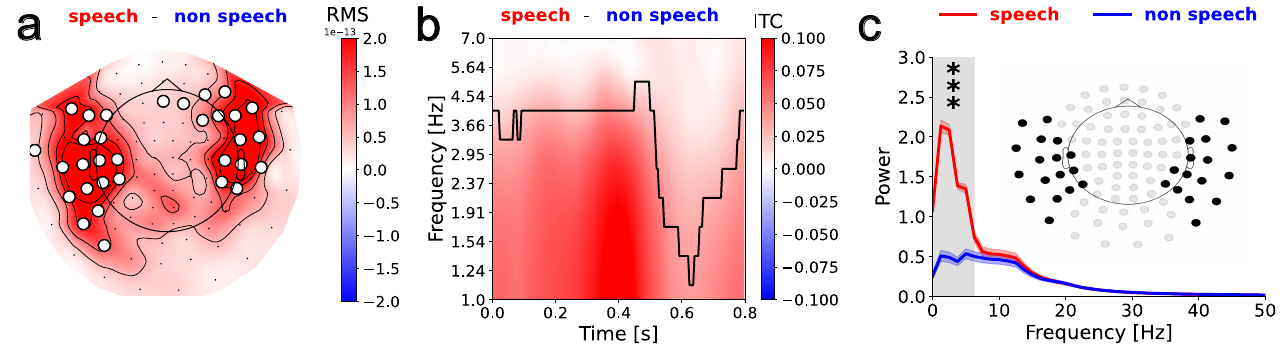}
    \caption{\textbf{MEG evoked response within subject}. Differences between speech and non-speech segments are shown as a proxy for auditory and linguistic processing. (\textbf{a}) Topographic map of the root mean square (RMS) difference between speech and non-speech evoked responses, averaged across sessions, showing stronger amplitude for speech over bilateral temporal sensors. Statistically significant sensors (p\textless0.001) are marked with white dots. (\textbf{b}) Time–frequency representation of the difference in inter-trial coherence (ITC) between speech and non-speech segments, averaged across sessions. Speech elicits greater phase consistency in low-frequency bands (\textasciitilde 2–7 Hz; delta–theta range). Statistically significant clusters (p\textless0.001) are outlined with a black contour. (\textbf{c}) Power spectral density (PSD) for speech (red) and non-speech (blue) segments, averaged across sessions, indicating increased low-frequency power during speech. Statistically significant frequency ranges (p\textless0.001) are highlighted with a semi-transparent grey region. For each panel, data are shown for a single representative participant (sub-0), averaged across recording sessions, where each session corresponds to one chapter (1-14) of \textit{A Study in Scarlet} from Sherlock Holmes audiobook.}
    \label{fig:sessig}
\end{figure}

\begin{figure}[H]
    \centering
    \includegraphics[width=0.98\linewidth]{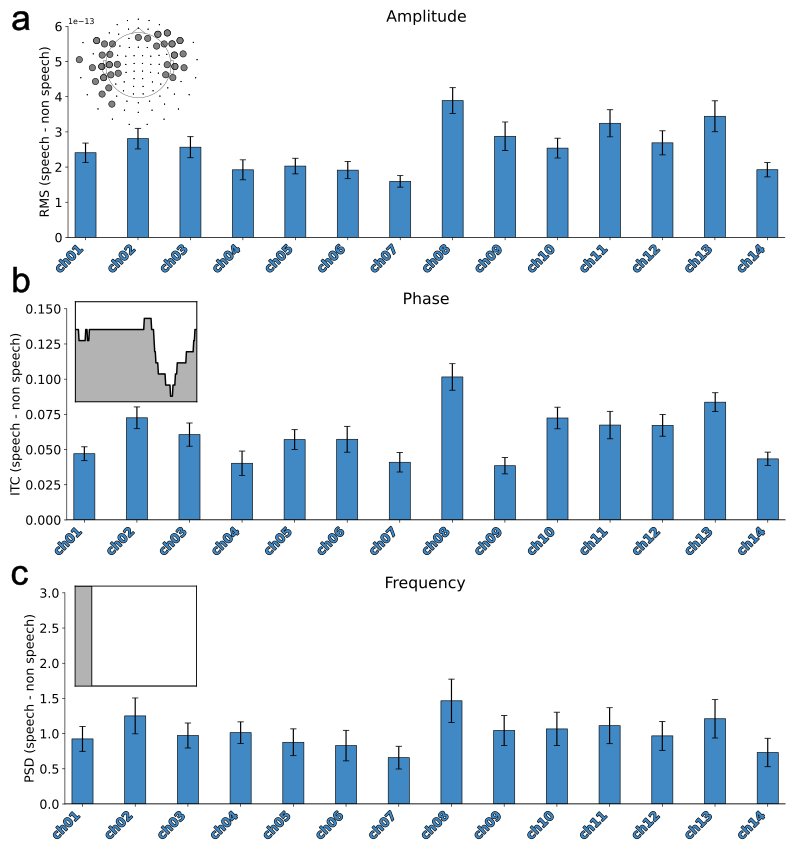}
    \caption{\textbf{Variability across recording sessions}. Spatial, temporal, and spectral components showing strong discrimination between speech and non-speech segments are illustrated across recording sessions to assess variability in amplitude, phase, and power. (\textbf{a}) Amplitude is quantified as the root mean square (RMS) of the evoked response. (\textbf{b}) Phase consistency is measured using inter-trial coherence (ITC). (\textbf{c}) Oscillatory power is measured using power spectral density (PSD). For each metric, the mean is computed within functionally relevant masks derived from statistical testing on aggregate data, shown as grey-shaded insets in each panel: a sensor mask for amplitude, a time–frequency mask for phase, and a frequency mask for power. Error bars indicate variability within the corresponding mask (across sensors, time–frequency points, or frequencies, depending on the panel). Data are shown for a single representative participant (sub-0), with each bar corresponding to one recording session, i.e., one chapter (1--14) of \textit{A Study in Scarlet} from Sherlock Holmes audiobook.}
    \label{fig:sesvar}
\end{figure}

We first assessed the consistency of these neural signatures within a single participant across multiple recording sessions (Figure~\ref{fig:sessig}). The evoked responses reveal clear commonalities: speech segments elicit stronger amplitude over bilateral temporal regions, increased phase alignment at low frequencies, and elevated low-frequency power for speech segments as compared to non-speech segments. These effects are robust at the group level and consistent with established findings in auditory neuroscience. However, when examining individual sessions in detail (Figure~\ref{fig:sesvar}), substantial variability emerges. The magnitude of amplitude differences, the degree of phase locking, and the extent of low-frequency power enhancement vary noticeably across sessions. Some sessions exhibit pronounced and clean separations between speech and non-speech, whereas others show attenuated or noisier patterns. This within-subject variability likely reflects fluctuations in attention, fatigue, and engagement with the narrative.

\begin{figure}[H]
    \centering
    \includegraphics[width=0.98\linewidth]{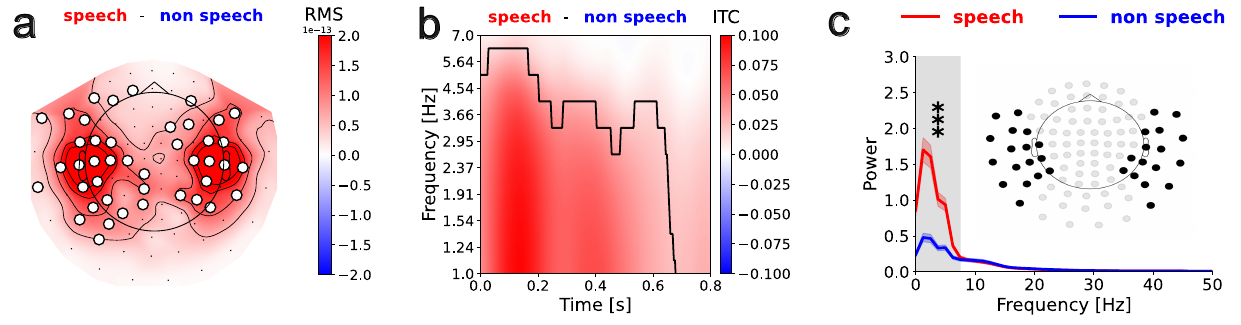}
    \caption{\textbf{MEG evoked response between subjects}. Data are shown for all participants (sub-0 to sub-32), averaged across subjects, for chapters 11–12 of \textit{Study in Scarlet} from Sherlock Holmes audiobook. Differences between speech and non-speech segments are shown as a proxy for auditory and linguistic processing. (\textbf{a}) Topographic map of the root mean square (RMS) difference between speech and non-speech evoked responses, averaged across subjects, showing stronger amplitude for speech over bilateral temporal sensors. Statistically significant sensors (p\textless0.001) are marked with white dots. (\textbf{b}) Time–frequency representation of the difference in inter-trial coherence (ITC) between speech and non-speech segments, averaged across subjects. Speech elicits greater phase consistency in low-frequency bands ($\sim$2–7 Hz; delta–theta range). Statistically significant clusters (p\textless0.001) are outlined with a black contour. (\textbf{c}) Power spectral density (PSD) for speech (red) and non-speech (blue) segments, averaged across subjects, indicating increased low-frequency power during speech. Statistically significant frequency ranges (p\textless0.001) are highlighted with a semi-transparent grey region.}
    \label{fig:megsig}
\end{figure}

\begin{figure}[H]
    \centering
    \includegraphics[width=0.98\linewidth]{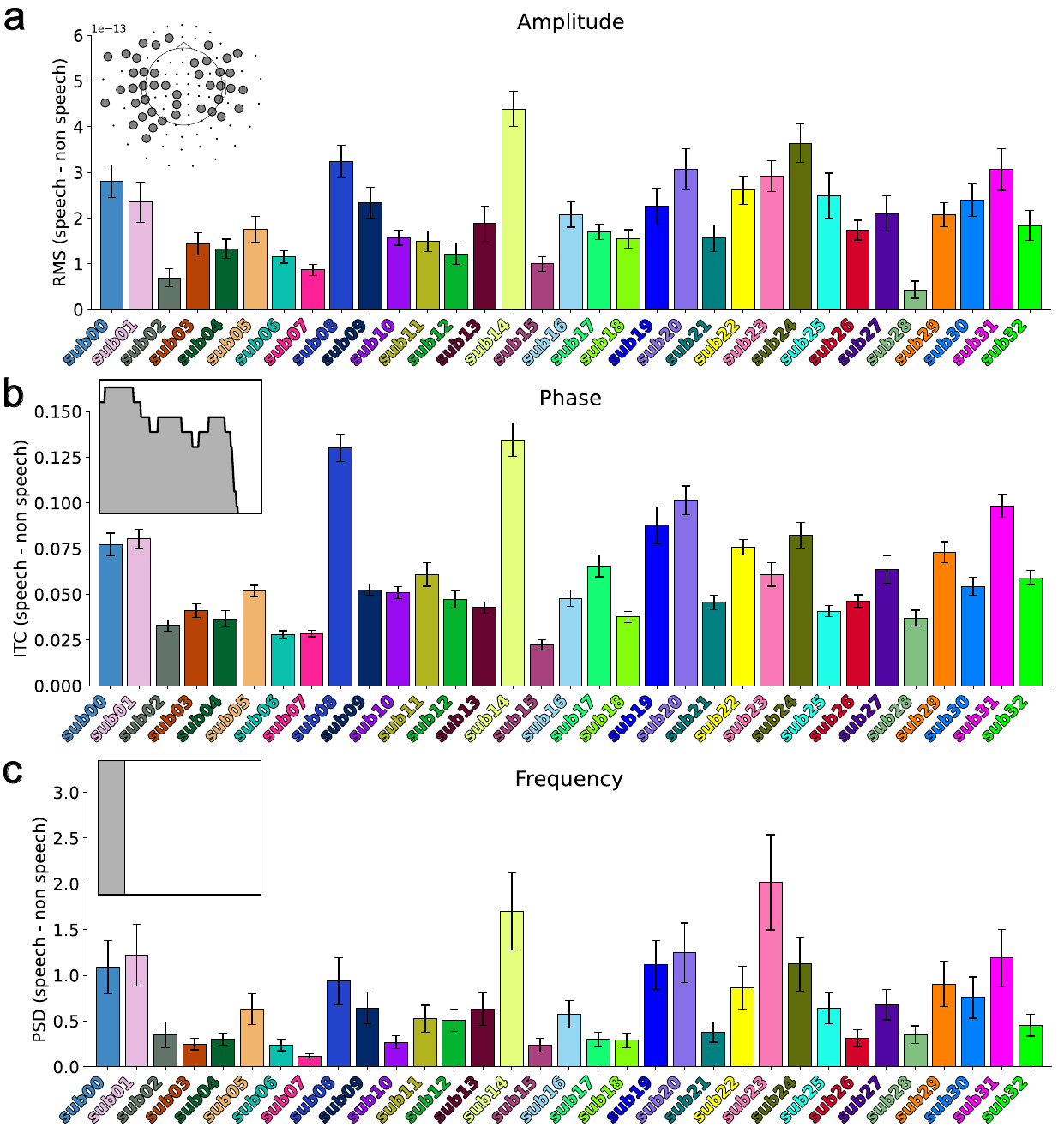}
    \caption{\textbf{Variability across subjects}. Spatial, temporal, and spectral components showing strong discrimination between speech and non-speech segments are illustrated across participants to assess variability in amplitude, phase, and power. (\textbf{a}) Amplitude is quantified as the root mean square (RMS) of the evoked response. (\textbf{b}) Phase consistency is measured using inter-trial coherence (ITC). (\textbf{c}) Oscillatory power is measured using power spectral density (PSD). For each metric, the mean is computed within functionally relevant masks derived from statistical testing on aggregate data, shown as grey-shaded insets in each panel: a sensor mask for amplitude, a time–frequency mask for phase, and a frequency mask for power. Error bars indicate variability within the corresponding mask (across sensors, time–frequency points, or frequencies, depending on the panel). Data are shown across participants (sub-0 to sub-32), with each bar corresponding to one subject. Colours differentiate participants.}
    \label{fig:megvar}
\end{figure}

A similar pattern is observed across participants (Figure~\ref{fig:megsig},~\ref{fig:megvar}), where shared neural signatures coexist with inter-individual variability. At the group level, the expected response patterns associated with speech perception are preserved: increased evoked amplitude over bilateral auditory cortices, stronger low-frequency phase coherence, and enhanced delta–theta power for speech relative to non-speech segments. Yet, the expression of these effects differs considerably between individuals. Some participants show strong and spatially focal responses, while others exhibit weaker or more diffuse patterns. These differences can arise from a combination of anatomical variability, differences in signal-to-noise ratio, age-related factors, and variability in cognitive engagement. Behavioural measures, such as responses to comprehension questions, further suggest that not all participants maintain the same level of attention throughout the recordings, which likely contributes to the observed heterogeneity in neural responses.

\vspace{-0.2cm}

Taken together, these analyses show that neural variability operates at multiple levels — both within and across subjects — and directly impacts the structure of the MEG signal used for decoding. Importantly, despite this variability, (Figures~\ref{fig:sessig} and ~\ref{fig:megsig}) reveal consistent and robust neural signatures across sessions and participants, including increased evoked amplitude over temporal regions, stronger low-frequency phase coherence, and enhanced delta–theta power during speech. These shared patterns indicate that a stable and meaningful signal is present across conditions, providing a common substrate that neural decoding models can exploit. At the same time, variability in the strength, spatial distribution, and reliability of these effects reflects fluctuations in attention, physiology, and recording conditions. As such, this variability is not merely noise to be eliminated, but a defining characteristic of the naturalistic speech perception task. It provides a critical testbed for evaluating the robustness of deep learning models, which must learn to extract stable and behaviourally relevant features while remaining invariant to these sources of variation. The dataset therefore enables two complementary forms of generalisation: within-subject generalisation across sessions and linguistic contexts, and between-subject generalisation across individuals. By explicitly incorporating both shared structure and variability, it offers a realistic and challenging benchmark for developing models that can scale to the complexity of naturalistic speech perception.

\section{Additional Decoding Experiments}
\label{ref:additional_decoding}

\subsection{Generalisation of Supervised and Pre-trained Models to Deep and Broad Data}

\begin{figure}[!htbp]
    \centering
    \includegraphics[width=0.9\linewidth]{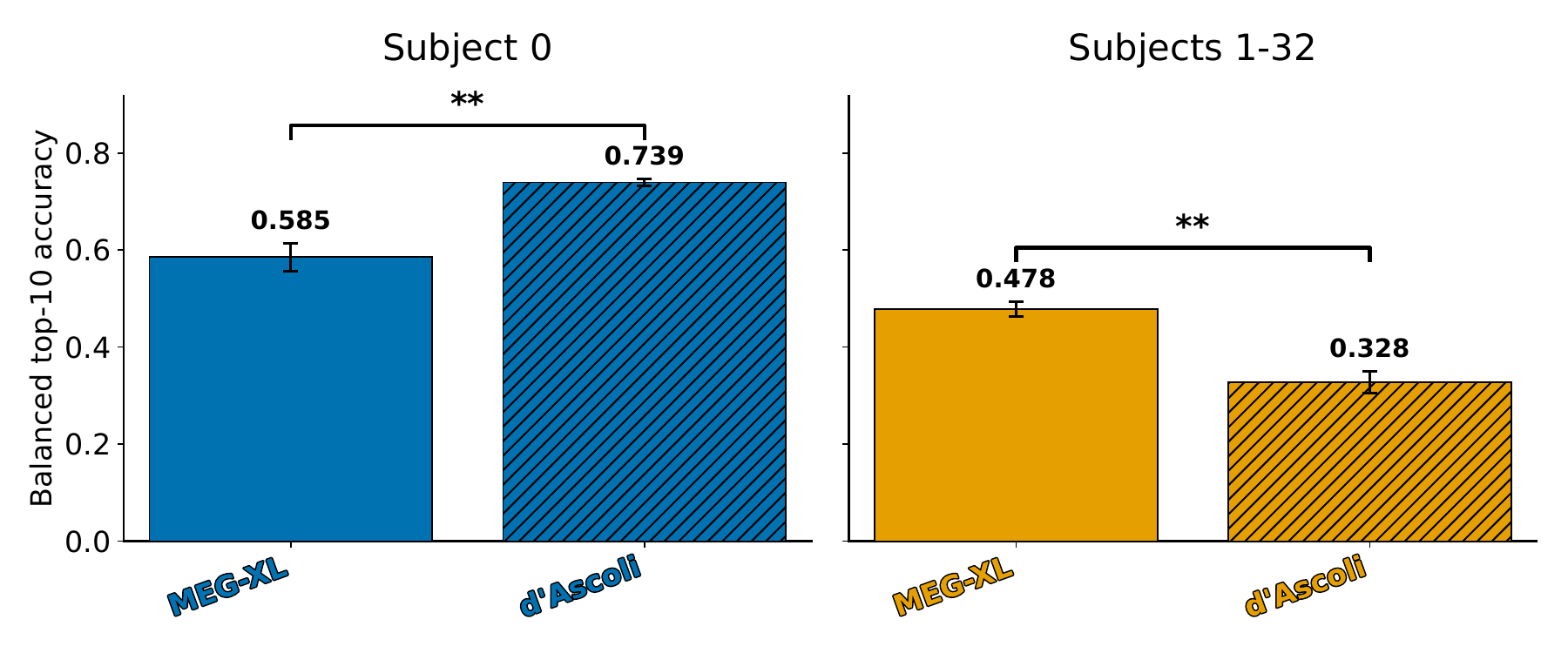}
    \caption{\textbf{d'Ascoli (Supervised) vs MEG-XL (Self-supervised).} We compare the supervised word decoding model proposed by \citet{dascoli2025natcomm} to fine-tuning the pre-trained self-supervised MEG-XL model proposed by \citet{jayalath2025meg-xl}. On deep Subject 0 data, the supervised model performs better, while on the shallow data of subjects 1-32, MEG-XL generalises better owing to its data-efficiency. Random chance is $0.2$. ** indicates $p<.01$ under a Mann-Whitney U-test.}
    \label{fig:megxldascoli}
\end{figure}

Building on the deep within-subject data of LibriBrain, LibriBrain100 introduces a new \textit{breadth} axis of scaling through 32 new subjects with relatively little within-subject data compared to Subject 0. These two regimes appear to favour different kinds of models, with supervised models performing best on the deep data portion (Subject 0), while fine-tuning a self-supervised model pre-trained over many subjects performs best for generalisation over Subjects 1-32 (Figure~\ref{fig:megxldascoli}). This reflects the findings in \citet{jayalath2025meg-xl}, who observe that with sufficient (deep) data, the statistical priors of pre-trained methods are superseded by rich in-domain data, whereas on shallow multi-subject data, the pre-trained multi-subject priors assist generalisation.

    \subsection{Measuring Information Transfer with LibriBrain100}
\label{app:info_transfer}

\begin{figure}[!htbp]
    \centering
    \includegraphics[width=0.7\linewidth]{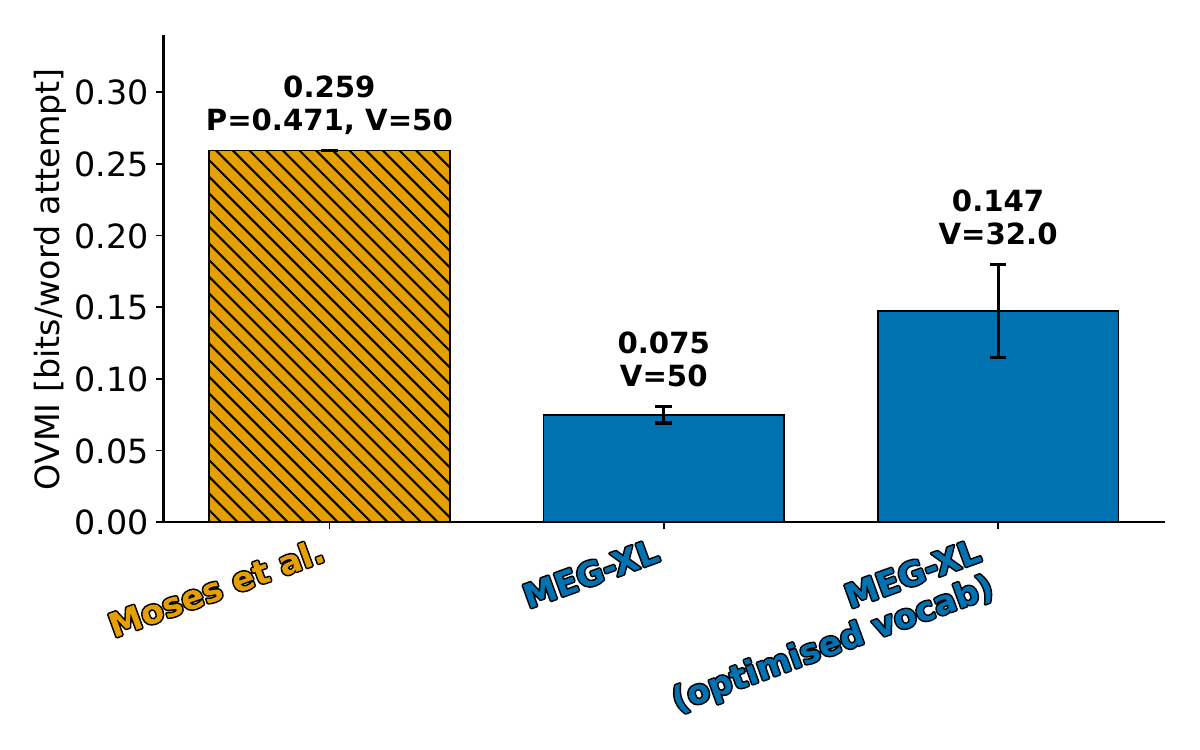}
    \caption{\textbf{Information transfer comparison.} We measure the OVMI achieved by \citet{moses2021nejm}, calculating the quantity from their reported $47.1\%$ decoder accuracy and 50-word vocabulary, against MEG-XL fine-tuned on LibriBrain with our target vocabulary and against an OVMI-optimised vocabulary. We use SUBTLEX-UK as the communication distribution $p$ that parameterises OVMI. Higher OVMI scores are better. The results here are provided as references for future benchmarks.}
    \label{fig:ovmi}
\end{figure}

The eventual hope in releasing LibriBrain100 is that it will accelerate progress towards non-invasive brain--computer interfaces that restore speech to paralysed patients. To this end, we benchmark the information throughput of models trained on LibriBrain100. We measure OVMI \citep{jayalath2026measuring}, an information-theoretic quantity for the mutual information between a user's intent and a decoding model, indexed to a particular communication distribution. We compare results on LibriBrain100 to that of a prior surgically implanted speech BCI used on a patient with anarthria \citep{moses2021nejm}. The early landmark result in \citet{moses2021nejm}, with a simple 50-word vocabulary, paved the way for significant advances in following years \citep{willett2023high, card2024nejm}. By measuring performance against \citet{moses2021nejm}, we aim to track how far the non-invasive decoding field is from reaching a similar breakthrough moment. The results in Figure~\ref{fig:ovmi} should be read with caution, however. LibriBrain100 is a heard speech dataset rather than an intended speech dataset, so the reported throughput does not represent that of a practical communication interface which would require evaluation against attempted or imagined speech decoding. Nevertheless, the results are indicative of significant progress being made towards strong performance on heard speech decoding through non-invasive methods.

\section{Additional Decoding Experiment Details}\label{sec:decoding_extras}

    \subsection{Computational Requirements}\label{app:compreq}

All experiments were performed on individual NVIDIA H100 GPUs on a system with 64 GiB of CPU memory. Fine-tuning MEG-XL with 100 hours of data took approximately 20 hours per run.

    \subsection{Target Words}\label{sec:target_words}

The 50-word vocabulary was designed to span both high-frequency content and function words. It includes pronouns (e.g., ``she'', ``him'', ``i'', ``we''), conjunctions and prepositions (e.g., ``and'', ``but'', ``on'', ``at''), determiners and quantifiers (e.g., ``the'', ``a'', ``any''), negation (e.g., ``not''), auxiliary and modal verbs (e.g., ``was'', ``is'', ``will'', ``can''), common lexical verbs (e.g., ``think'', ``do''), and a small number of content words (e.g., ``people'', ``time'', ``good'', and ``new'') (Table~\ref{tab:words}).

Even with only 50 words, the vocabulary can support a range of short, practically useful utterances in the context of assistive communication. Examples include state reports (``i am good'', ``i am not good'', ``i think this is good'', ``i can do that'', ``i am out of it''), requests (``can i have that'', ``can he be there'', ``do not do that''), location or attention cues (``i will be there'', ``is it time''), and social reference (``she was good to me'').

\begin{table*}[!htbp]
\centering
\small
\setlength{\tabcolsep}{15pt}
\begin{tabular}{rl@{\hspace{0.8em}}rl@{\hspace{0.8em}}rl@{\hspace{0.8em}}rl@{\hspace{0.8em}}rl}
\toprule
\# & Word & \# & Word & \# & Word & \# & Word & \# & Word \\
\midrule
1  & is       & 11 & he       & 21 & but      & 31 & at       & 41 & do \\
2  & the      & 12 & that     & 22 & will     & 32 & out      & 42 & can \\
3  & a        & 13 & have     & 23 & so       & 33 & our      & 43 & time \\
4  & to       & 14 & this     & 24 & all      & 34 & am       & 44 & think \\
5  & it       & 15 & they     & 25 & my       & 35 & it’s     & 45 & good \\
6  & i        & 16 & of       & 26 & for      & 36 & had      & 46 & always \\
7  & not      & 17 & there    & 27 & she      & 37 & him      & 47 & new \\
8  & was      & 18 & and      & 28 & were     & 38 & an       & 48 & people \\
9  & we       & 19 & are      & 29 & any      & 39 & very     & 49 & as \\
10 & be       & 20 & in       & 30 & really   & 40 & has      & 50 & on \\
\bottomrule
\end{tabular}
\caption{\textbf{The 50-word vocabulary used in the word classification task}.}
\label{tab:words}
\end{table*}

\begin{figure}[H]
    \centering
    \includegraphics[width=1.0\linewidth]{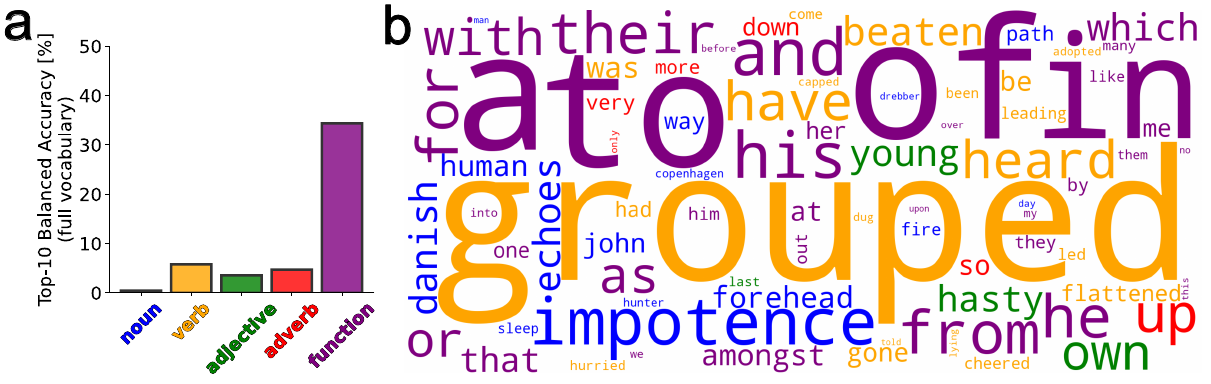}
    \caption{\textbf{Model predictions on full vocabulary}. (\textbf{a}) Top-10 accuracy across all word occurrences in Chapters 11 and 12 of A Study in Scarlet and across all 32 subjects, grouped by Part-of-Speech (POS) category (purple function, red adverb, green adjective, orange verb, blue noun). Predictions were evaluated against the full vocabulary, and accuracy is defined as the proportion of events for which the true word appears within the top-10 predicted words. (\textbf{b}) Word cloud of the 100 best-predicted words, ranked by top-10 balanced accuracy. Word size is proportional to top-10 balanced accuracy, and word colours indicate their corresponding Part-of-Speech categories.}
    \label{fig:wordpred}
\end{figure}

\section{Ethical Considerations}\label{sec:ethics}

The development and release of the LibriBrain100 dataset raise several important ethical considerations, which we have addressed throughout the research process:

\textbf{Informed consent and participants privacy}. All participants provided informed consent for data collection and explicitly approved the sharing of pseudoanonymised data for research purposes, in accordance with the University of Oxford’s ethical oversight procedures. To safeguard privacy, all released data are subject to anonymisation procedures, including the removal or replacement of any information that could directly or indirectly identify individual participants. These steps are designed to minimise re-identification risk while preserving the utility of the dataset.

\textbf{Open science and reproducibility}. We intentionally rely where possible on public-domain materials (e.g., LibriVox audiobooks and Project Gutenberg texts), and use established speech and podcast corpora subject to their respective terms and licences (TIMIT, MOCHA-TIMIT, and The Moth). This commitment to openness extends to our open-source tooling and publicly released data, which are documented to support reproducibility.

\textbf{Dual-use considerations}. While the primary motivation of this work is to support assistive communication technologies for clinical populations, brain decoding methods could be misused in ways that compromise privacy if applied without consent. Although current non-invasive approaches remain far from enabling such uses, we stress the importance of maintaining strong ethical standards, including informed consent and respect for participant autonomy. By releasing data and methods, we aim to support the development of shared ethical guidelines and responsible practices within the research community.

\textbf{Long-term data stewardship}. We are committed to ensuring the long-term availability and integrity of the dataset. The BIDS-formatted version will be hosted on established public platforms (e.g., OSF), and the dataset is also distributed via Hugging Face. Together with code hosted on GitHub, this provides redundancy and helps ensure sustained accessibility.

\end{document}